\documentclass[sigconf]{acmart}
\usepackage{subfigure}
\usepackage{bm}
\usepackage{ulem}

\AtBeginDocument{%
  \providecommand\BibTeX{{%
    \normalfont B\kern-0.5em{\scshape i\kern-0.25em b}\kern-0.8em\TeX}}}

\copyrightyear{2021} 
\acmYear{2021} 
\setcopyright{acmcopyright}\acmConference[MM '21]{Proceedings of the 29th ACM International Conference on Multimedia}{October 20--24, 2021}{Virtual Event, China}
\acmBooktitle{Proceedings of the 29th ACM International Conference on Multimedia (MM '21), October 20--24, 2021, Virtual Event, China}
\acmPrice{15.00}
\acmDOI{10.1145/3474085.3475334}
\acmISBN{978-1-4503-8651-7/21/10}

\definecolor{mygreen}{rgb}{0.25,0.87,0.81}

\begin{document}
 \fancyhead{}
\title{Parametric Reshaping of Portraits in Videos}

\author{Xiangjun Tang}
\affiliation{%
  \institution{{State Key Lab of CAD\&CG, Zhejiang University}} \country{}}
\email{fcsx1tf@163.com}

\author{Wenxin Sun}
\affiliation{%
  \institution{{State Key Lab of CAD\&CG, Zhejiang University}} \country{}}
\email{japonica99@163.com}

\author{Yong-Liang Yang}
\affiliation{%
  \institution{{University of Bath}}\country{}}
\email{y.yang@cs.bath.ac.uk}

\author{Xiaogang Jin}
\authornote{Corresponding author.}
\affiliation{%
  \institution{{State Key Lab of CAD\&CG, Zhejiang University}}
  \country{} }
\email{jin@cad.zju.edu.cn}
\renewcommand{\shortauthors}{Trovato and Tobin, et al.}

\begin{abstract}
Sharing short personalized videos to various social media networks has become quite popular in recent years. This raises the need for digital retouching of portraits in videos. However, applying portrait image editing directly on portrait video frames cannot generate smooth and stable video sequences. To this end, we present a robust and easy-to-use parametric method to reshape the portrait in a video to produce smooth retouched results. Given an input portrait video, our method consists of two main stages: stabilized face reconstruction, and continuous  video reshaping. In the first stage, we start by estimating face rigid pose transformations across video frames. Then we jointly optimize multiple frames to reconstruct an accurate face identity, followed by recovering face expressions over the entire video. 
In the second stage, we first reshape the reconstructed 3D face using a parametric reshaping model reflecting the weight change of the face, and then utilize the reshaped 3D face to guide the warping of video frames. We develop a novel signed distance function based dense mapping method for the warping between face contours before and after reshaping, resulting in stable warped video frames with minimum distortions. In addition, we use the 3D structure of the face to correct the dense mapping to achieve temporal consistency. 
We generate the final result by minimizing the background distortion through optimizing a content-aware warping mesh.
Extensive experiments show that our method is able to create visually pleasing results by adjusting a simple reshaping parameter, which facilitates portrait video editing for social media and visual effects.

\end{abstract}


\begin{teaserfigure}
\centering
  \includegraphics[width=\textwidth]{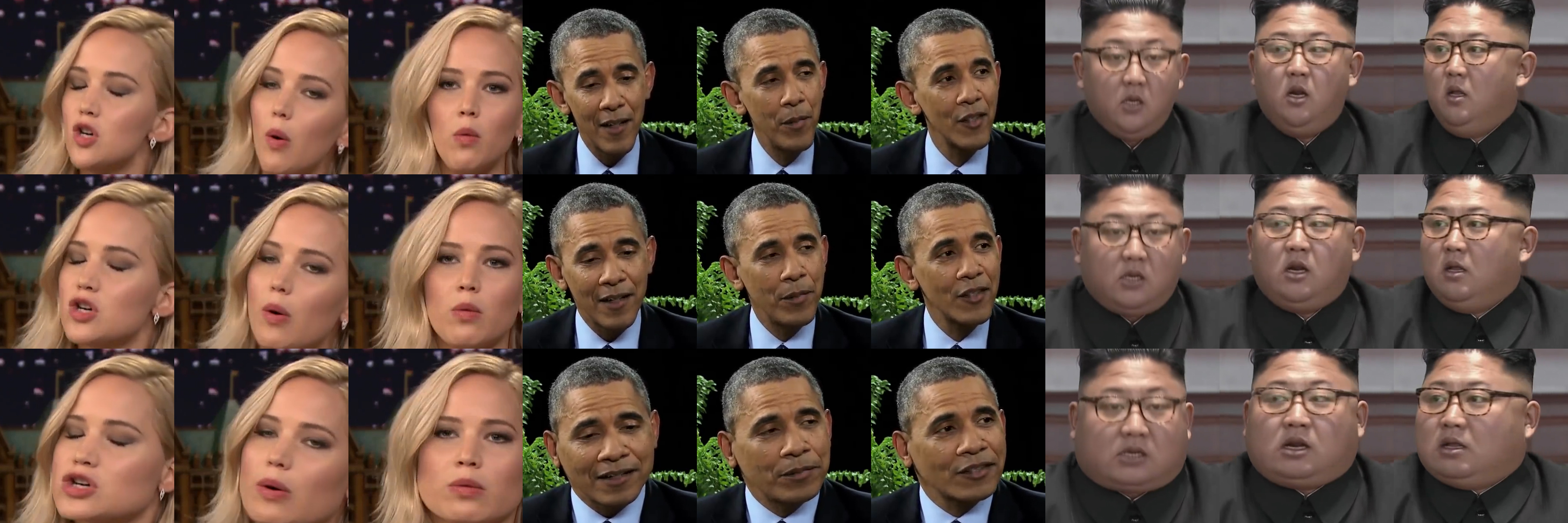}
  \caption{Our parametric reshaping method allows users to reshape the portrait in a video footage easily by simply adjusting a reshaping parameter. Given an input portrait video sequence (second row), our approach can reshape the portrait in the video with weight-change such that the face appears thinner (the first row) or rounder (the third row), respectively.}
  \Description{}
  \label{fig:teaser}
\end{teaserfigure}

\maketitle
\section{Introduction}

Due to the recent development of social networks and personalized media, more and more people have become active to share their own photos and videos to others using mobile phones. Portrait editing techniques are often utilized to create special portrait effects through face stretch, exaggeration, beautification, etc., such that the resultant portraits are more appealing.
Research attention has also been focused on different face retouching methods to edit face colors, textures, styles, and even shapes for portrait images \cite{leyvand2008data} \cite{shih10.1145/2601097.2601137} \cite{Zhao2018Portrait}  \cite{XiaoPortraits}.
%
Compared with portrait images, portrait video editing has been much less explored. Although video can be directly processed by applying image editing to each individual video frame, this can easily lead to various artifacts due to the lack of editing consistency and temporal coherence across neighboring video frames.

Our aim is to generate high-quality portrait video reshaping results (see Figure~\ref{fig:teaser}) by editing the overall shape of the portrait faces according to natural face deformation in real world. This can be used for applications such as shapely face generation for beatification, and face exaggeration for visual effects. Compared with portrait image reshaping~\cite{Zhao2018Portrait}\cite{XiaoPortraits}, we also employ a 3D face reconstruction based approach to guide 2D portrait editing with faithful face deformation in 3D. However, for plausible video reshaping, a key difference in our setting is that our new problem requires the \textit{consistency} and \textit{coherency} of the reshaped portraits over the entire video.
This new requirement poses two challenges in practice, i.e., how to achieve consistency and coherency not only on the reconstructed 3D faces, but also on the reshaped video frames.

In this paper, we present a parametric reshaping method that can generate quality reshaped portrait videos. Given an input video, our method contains two main stages to address the above two challenges. The first stage mainly focuses on consistent and coherent face reconstruction. Based on a state-of-the-art face parametric model, we employ a multi-phase optimization that robustly estimates pose transformations, face identity parameters, and face expression parameters, respectively. 
%
Our method particularly concerns the face contour's stability in face tracking to avoid artifacts after reshaping. To estimate a consistent face identity across all frames, we jointly optimize the representative frames rather than all frames to reduce computational cost.
The energy terms in the optimization are carefully chosen through extensive ablation study by taking both spatial-temporal coherence and computational efficiency into consideration.
%
The second stage attempts to generate reshaped video frames without visual artifacts.
Guided by reshaped faces in 3D, we employ a content-aware image warping method to deform each frame. To avoid warping artifacts caused by face occlusions after reshaping, we use a SDF to construct the dense 2D mapping of the face contour before and after reshaping. In addition, we use the 3D structure of the face to correct the dense mapping. The dense 2D mapping is then transferred to a sparse set of grid points to warp video frames.

We evaluate our work on a variety of videos with different portrait characteristics including gender, face color, hair style, pose variations, etc. The results in the paper and supplemental video both demonstrate the effectiveness and robustness of our work. Extensive comparisons with different design choices also verify the superiority of our method.

The major contributions of this work are summarized as the following:
\begin{itemize}
\item We present the first robust parametric reshaping approach for high-quality reshaped portrait video generation.
\item We propose an efficient and stable 3D face reconstruction method using a multi-phase optimization strategy with a refined dense flow energy.
\item {We propose an effective correspondence estimation method based on signed distance function and 3D information to deform a portrait video without artifacts.}
\end{itemize}

\section{ Related work }

In this section, we discuss previous approaches that are relevant to our method. As there is no prior work on parametric reshaping of portraits in videos, we organize this section according to the two stages of our method by reviewing related works on video-based reconstruction and video deformation, respectively.
\subsection{Video based face reconstruction}

\textbf{Morphable Model.} 
Monocular 3D face reconstruction is an ill-posed optimization problem, which requires priors of face identity and expression. 
Blanz and Vetter \shortcite{BlandVetter99} proposed the 3D Morphable Model (3DMM) using principal component analysis (PCA) on 3D scans, which can be used as identity priors for reconstruction. Blendshapes Model \cite{lewis2014practice} offered expression priors using faces with the same topology but different expressions. Over the past years, a variety of works used both linear models and their extensions for face reconstruction \cite{Egger2020}. Surrey Face Model proposed by Huber et al.~\shortcite{Huber2015} is a multi-resolution 3DMM that contains different mesh resolution levels. Booth et al. \shortcite{Booth} extended 3DMM by combining a statistical model of face shape with an ``in-the-wild'' texture model to reduce illumination parameters in optimization. Rapid progress has been made for improving reconstruction speed, accuracy, and ease of use even in unconstrained conditions \cite{Thies2018}. However, the type and amount of training data constrained the performance of linear 3DMM. Tran and Liu \shortcite{Tran} proposed to learn a nonlinear 3DMM from face images rather than 3D face scans to produce a more extensive database. Tran et al. \shortcite{Trana} further improved the nonlinear 3DMM in both learning objective and network architecture to achieve high-fidelity face reconstruction results. {Based on a set of 4,000 high resolution facial scans, Li et al. \shortcite{li_learning_2020} proposed a deep-learning based morphable face model.}


\textbf{Video based reconstruction.}
Although portrait video contains abundant frames, the joint optimization of face pose, identity, and expression is still challenging. Simply adding more constraints to the optimization is hard to achieve satisfactory results. 
Thies et al. \shortcite{thies2016face2face} used model-based non-rigid bundle adjustment over keyframes with different head poses. Cao et al. \shortcite{Cao2018} proposed the on-the-fly method for face tracking using a dynamic rigidity prior learned from realistic datasets.
This method can achieve plausible results when the landmarks are stable and mostly visible. However, the reconstruction of the current frame depends on the result of the previous frame. If the landmarks are not accurate enough or very different between frames, it is still challenging to achieve accurate and continuous results.
%

%

\begin{figure*}
	\centering
	\includegraphics[width=.99\textwidth]{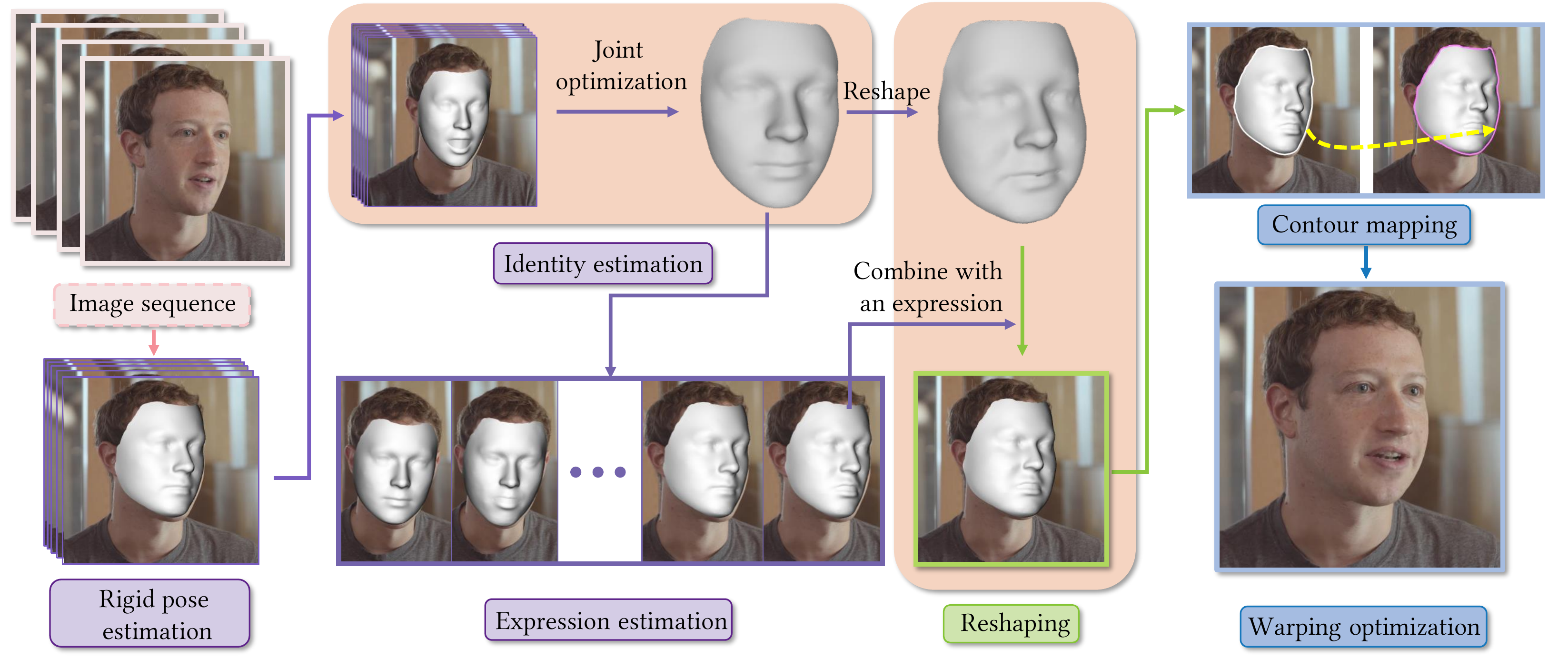}
	\caption{Given an image sequence, we first estimate the rigid  pose of the face for each frame, which is crucial for computing a consistent identity across all frames. Then, we leverage an optimization method by using the most representative frames jointly to estimate the whole sequence's consistent identity parameters. Next, we estimate each frame's expression parameters. Given a reshaping parameter, a linear regression method is employed to generate the reshaped 3D face model sequence by combining the identity and the expression parameters. After that, we employ an SDF based method to construct the 2D dense mapping of the face contour before and after reshaping, and perform a content-aware warping optimization to deform the video in order to get our final result. }
	\label{fig:Pipeline}
\end{figure*}

\subsection{Portrait video deformation}
For reducing image distortion after face editing, content-aware image warping becomes a powerful tool in a wide range of editing applications   \cite{Kaufmann2013} \cite{shih2019distortion} \cite{XiaoPortraits}.
However, it is challenging to generate continuous and stable deformations across all video frames because the mappings from source to target of consecutive frames are usually inconsistent. Chen et al.~\shortcite{Chen2013} noticed the importance of a consistent blending boundary and presented a video blending approach to merge the gradient of source and target videos.
Thies et al. \shortcite{thies2015expression_transfer} transferred expressions by mapping the expression parameters from the source to the target without changing other parameters, such that the target identity, rigid head motion, and scene lighting can be preserved. {Moreover, portrait video attributes such as head pose, facial expression, scene illumination can be manipulated by GAN-based methods \cite{tewari_pie_2020} \cite{nagano_pagan_2019} \cite{ZengTalking} \cite{ChenSimSwap}.}
%
%

\section{Overview}

This work proposes a novel method that parametrically reshapes a portrait video, making the resultant portrait sequence realistic and stable. This requires robust extraction of portrait shapes from the video and consistent deformation of video frames to reshape the portraits, which are all addressed by our method.

Figure \ref{fig:Pipeline} shows the pipeline of our method. Given a portrait image sequence, our method consists of two main stages. In the first stage (Section \ref{section:reconstruction}), we utilize a video-based face reconstruction approach to faithfully reconstruct high-quality face identity with stable poses and expressions. We first estimate the head pose in each frame, which is crucial to estimate a consistent identity across frames. Then we find the $k$ frames that best represent the face identity and jointly optimize a consistent face identity. Finally we estimate the face expression of each frame to achieve the whole 3D face sequence.
In the second stage (Section~\ref{section:reshaping}), we generate reshaped faces in 3D based on the reconstruction results, then leverage deformed faces in 3D to guide the reshaped portrait video generation in 2D. We first reshape the reconstructed 3D neutral face model and combine it with facial expression and pose for each video frame. 
%
%
We then employ an SDF based method to construct a 2D dense mapping of the face contour before and after reshaping. A content-aware warping optimization is used to deform portrait frames according to reshaped faces in 3D, resulting in the final reshaped portrait video.

\section{Video based face reconstruction }\label{section:reconstruction}




In this section, we first describe the parametric face model and the objectives that our optimization is based on. Then we elaborate how we optimize face pose, identity, and expression step by step while taking into account robustness and efficiency.

\subsection{Parametric Face Model and Objectives }
\label{subsec:facemodel}
{
The parametric face model can be represented via a linear combination of identity basis vectors and expression basis vectors:

\begin{equation}
     \bold{s}=\bold{a}_s+\sum_{i=1}^{m_s} \alpha_i\bold{b}_i^s+\sum_{i=1}^{m_e} \beta_i \bold{b}_i^e,
\end{equation}
where $\bold{a}_s\in R^{3n}$ is the mean face identity ($n$ is the number of vertices of the face), $\bold{b}_i^s\in R^{3n}$ is one of the $m_s$ identity basis vectors, and $\bold{b}_i^e \in R^{3n}$ is one of the $m_e$ expression basis vectors. $\bm{\alpha}=[\alpha_1,...,\alpha_{m_s}]$,  $\bm{\beta}=[\beta_1,...,\beta_{m_e}]$ are the identity coefficients and expression coefficients, respectively.}

{The parametric face model proposed by Huber et al.~\shortcite{Huber2015}} consists of $m_s= 63$ facial identity coefficients and $m_e=6$ expression coefficients. Although this model does not express ample expressions, it is capable of representing common portrait faces.

We denote the projection operator by $\Pi :R^3\to R^2$ , which maps the $k$-th mesh vertex to image coordinate $\bold{p}_i \in R^2$ as:
\begin{equation}
    \bold{p}_i=\Pi(\bold{r},\bold{t},\bm{\alpha},\bm{\beta})_k,
\end{equation}
where $\bold{r}\in R^{3}$ and $\bold{t}\in R^{3}$ are the rotation and translation parameters of the face model, respectively.
Now we present our objective energy terms based on the aforementioned parameters which will be used in the latter optimization.

\textit{Landmark energy.} We denote a set of 2D facial landmark indices as $\mathcal{L}=\{L_1,...,L_{NL}\}$ ($NL$ is the number of landmarks). The first objective aims to match all 3D landmarks to their corresponding 2D landmarks after projection (assume 2D and 3D landmark indices are the same for simplicity):
\begin{equation}
E_{land}=||\Pi(\bold{r},\bold{t},\bm{\alpha},\bm{\beta})_i-\bold{p}_i||^2_{\forall i\in \mathcal{L}}.
\label{eq:land}
\end{equation}

\textit{Contour energy.} The part of face away from the camera is likely to have offset landmarks in 2D due to the incomplete face region induced by occlusion (see an example in Figure \ref{fig:contour}(a) near the eye). 
Therefore we also define a contour energy to match face boundary between 3D and 2D.
%
Note that the boundary of the projected 3D face model will change when the pose changes, and thus it is impossible to obtain the accurate boundary. However, as our method estimates face rigid pose first (see Section~\ref{sec:facetracking}), we can utilize the boundary of the approximate 3D face model obtained from the rigid pose estimation step to constrain the alignment.


Then the contour energy can be expressed as follows:
\begin{equation}
E_{contour}=||\Pi(\bold{r},\bold{t},\bm{\alpha},\bm{\beta})_k-\bold{p}_i||^2_{\forall i\in \mathcal{L}_b},
\end{equation}
{where $\mathcal{L}_b$ is the set of indices of contour landmarks, and $k$ is the corresponding vertex's index on the 3D face boundary.}

\textit{Alignment energy.} By now, we can define the \textit{alignment energy} as the linear combination of landmark energy and contour energy:
\begin{equation}
    E_{align}=E_{land}+\sigma E_{contour}.
    \label{eq:align}
\end{equation}
Empirically, we set $\sigma=0.5$.

\textit{Prior energy.} A regularization term is also defined to constrain the reconstructed face to be regular in face space:
\begin{equation}
   E_{prior}=w_{prior}(||\bm{\alpha}||_2+||\bm{\beta}||_2),
\end{equation}
 where $w_{prior}$ is the weight of the regularization. We empirically set $w_{prior}$ as 0.4.
 
\textit{Temporal coherence energy.} 
Typically the temporal coherence energy reduces the parameter difference between previous frame and current frame. The expression coherence term can be defined as:
\begin{equation}
E_{temp}^{expr}=||\bm{\beta}-\bm{\beta}'||^2.
\label{eq:exprT}
\end{equation}
Similarly, the pose coherence term can be defined as:
\begin{equation}
E_{temp}^{pose}=||\bold{r}-\bold{r}'||^2+\gamma||\bold{t}-\bold{t}'||^2,
\label{eq:poseT}
\end{equation}
where $\gamma$ is the parameter to balance the effect of translation and rotation. In our experiment, we set $\gamma$ as the reciprocal of face length.
Then the overall temporal coherence energy is defined as:
\begin{equation}
    E_{temp}=E_{temp}^{pose}+\sigma E_{temp}^{expr},
\end{equation}
where we set $\sigma=2$ in our experiments.

\begin{figure}[t]
	\centering
	\subfigure[Input portait image]{
	\includegraphics[width=0.33\linewidth]{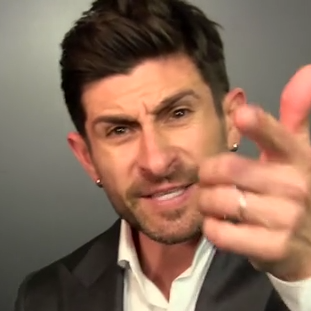}
	}%
    \subfigure[Using optical flow only]{
        \includegraphics[width=0.33\linewidth]{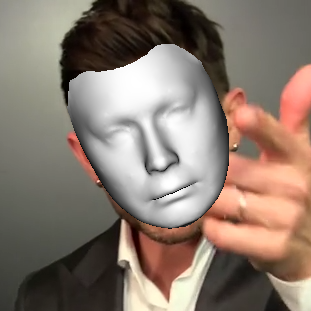}
    }%
    \subfigure[Our result]{
        \includegraphics[width=0.33\linewidth]{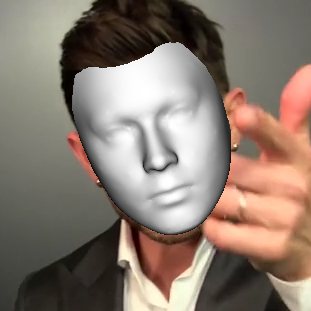}
    }%

	\caption{Inaccurate optical flow map due to occluded face (a) may lead to failed results (b) by simply applying Eqn. \ref{equ:opticalEnergy}. (c) shows our correct result.}
	\label{fig:opticalFailure}
\end{figure}

\textit{Dense flow energy.} 
Inspired by Cao et al. \shortcite{Cao2018}, we use optical flow to construct dense correspondences to overcome landmark detection errors.
With pose parameter $\bold{t}', \bold{r}'$ and expression coefficients $\bm{\beta}'$ from the previous frame, the dense flow energy is defined as:
\begin{equation}
\label{equ:opticalEnergy}
E_{optic}=||\Pi(\bold{r},\bold{t},\bm{\alpha},\bm{\beta})_i-\Pi(\bold{r}',\bold{t}',\bm{\alpha},\bm{\beta}')_i - \bold{U}_i||^2_{\forall{i\in \mathcal{L}_b}},
\end{equation}
{where $\bold{U}_i$ is the motion vector.}
{As expression change is not our main concern, }
we define the energy term only on face contour $\mathcal{L}_b$ rather than the full face, which also reduces computational cost
without affecting visual quality. 
However, artifacts occur 
when the optical flow map is inaccurate on face contour due to occlusions caused by other objects (see Figure \ref{fig:opticalFailure} for a failure case). { The change of optical flow on the face contour is mainly induced by camera movement and face pose transformations. Besides, the occluded regions may have different values on the flow map. Employing a low-frequency filter to eliminate the outliers can preserve the correct part. }


\subsection{Stabilized Face Tracking}\label{sec:facetracking}

Reconstructing face shape on the fly is convenient for real-time applications. However, it is hard to ensure that the identity reconstructed from the first frame is the same as that from the last frame. In contrast, joint optimization for all frames can achieve a consistent face identity but is too costly. As such, we first estimate the face pose for each frame, and then choose the consecutive $k$ frames which best represent the face identity coefficients. After that, we apply joint optimization for these $k$ frames to achieve the accurate face identity. Finally we estimate the expression coefficients of each frame to obtain the final reconstruction result.
The advantage of approximating face pose first is that face poses provide stable projected face contours and face regions in 2D, which enables us to employ the contour energy and revised dense flow energy defined in Section \ref{subsec:facemodel} for identity and expression reconstruction.


\subsubsection{Rigid Pose Estimation}
Based on the objectives defined in Section \ref{subsec:facemodel}, the pose estimation energy is defined as:
\begin{equation}
    E_{pose}=\lambda_{land}E_{land}+\lambda_{temp}^{pose}E_{temp}^{pose}+\lambda_{optic}E_{optic},
\end{equation}
where we empirically set $\lambda_{land}=0.6$, $\lambda_{temp}^{pose}=0.9$, and $\lambda_{optic}=0.5$. At this stage, we fix the identity parameters and expression parameters.
{Our extensive experiments show that such a strategy enables us to achieve  more stable final results as the energy terms involving all variables may conflict with each other during the optimization.}

\subsubsection{Identity Estimation}

It is hard to estimate the full shape of a face if only partial face is visible in the image. After pose estimation, we choose the consecutive $k$ $(k\leq 10)$ frames whose faces are facing to the camera, and apply bundle optimization \cite{triggs1999bundle} to these frames.







Based on the above and the objectives defined in Section~\ref{subsec:facemodel}, the identity estimation energy is defined as:
\begin{equation}
\begin{aligned}
E_{identity}=&\sum_i^{k+i}[\lambda_{align}(E_{align})_i+\lambda_{optic}(E_{optic})_i+\\&\lambda_{temp}(E_{temp})_i+(E_{prior})_i].
\end{aligned}
\end{equation}
To estimate the face identity as accurate as possible, we intentionally increase the weight of the landmark energy term $E_{align}$.
In our experiments, we set $\lambda_{align}=0.7$, $\lambda_{optic}=0.1$ and  $\lambda_{temp}=0.2$.


\subsubsection{Expression Estimation}

After shape estimation, we solve for face expressions by fixing shape coefficients. Similar to $E_{shape}$, we can define the expression energy as follows:
\begin{equation}
\begin{aligned}
E_{expr}&=\lambda_{align}(E_{align})_i+\lambda_{optic}(E_{optic})_i+\\&\lambda_{temp}(E_{temp})_i+(E_{prior})_i,
\end{aligned}
\end{equation}
where we empirically set $\lambda_{align}=0.9$, $\lambda_{optic}=0.5$ and $\lambda_{temp}=0.5$. However, we apply it for all video frames.

\section{Reshaping}\label{section:reshaping}

In this section, we reshape the 3D faces reconstructed in the last section and use them to guide the generation of the reshaped portrait video. More specifically, given a reshaping parameter, we first use a linear regression model to generate the reshaped faces in each frame (Section \ref{subsec:facereshape}). Then we employ image warping to deform each frame respecting the corresponding reshaped face (Section \ref{subsec:videodeform}).

\subsection{3D Face Reshaping}
\label{subsec:facereshape}

Xiao et al. \shortcite{XiaoPortraits} proposed a reshaping model to generate reshaped portrait images based on an estimated adjusting parameter. We extend this method from reshaping one monocular image to reshaping the whole image sequence. 
{The reshaping model deforms the whole face by utilizing a scalar parameter $\delta$ as input.}
We denote the reshape operator as $f(\bm{X};\delta)$.



{
For a sequence of reconstructed 3D faces $\bm{X}_i(\bm{\alpha},\bm{\beta}^{(i)})$, where $\bm{\beta}^{(i)}$ is the expression coefficients vector of the $i$-th frame, the reshaped 3D face model $\bm{X}^{*}_i$ is defined as the linear combination of the neutral reshaped face model and the expression coefficients $\bm{\beta}^{(i)}$:
\begin{equation}
    \bm{X}_i^*=f(\bm{X}(\bm{\alpha},\bm{\beta}_0);\delta)+\sum_{k=1}^{m_e} \beta^{(i)}_k\bold{b}_k^e,
\end{equation}
where $\bm{\beta}_0$ is the expression coefficients vector of the neutral face. 
}

\subsection{Consistent Video Deformation}
\label{subsec:videodeform}

\begin{figure}[b]
	\centering
	\subfigure[Original grid]{
	\includegraphics[width=0.33\linewidth]{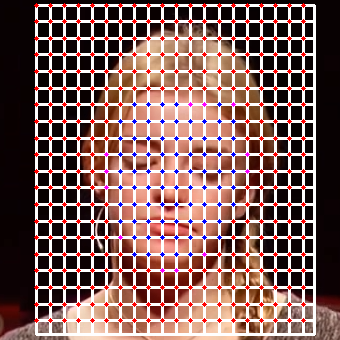}
	}%
    \subfigure[Grid after MLS]{
        \includegraphics[width=0.33\linewidth]{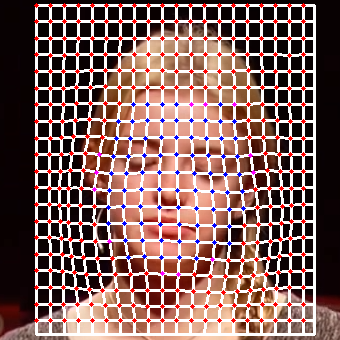}
    }%
    \subfigure[Grid after optimization]{
        \includegraphics[width=0.33\linewidth]{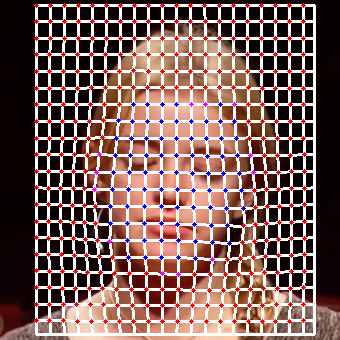}
    }%

	\caption{Grid deformation through MLS and optimization. (a) The original uniform grid $M_u$. (b) The grid after MLS based  reshaping. (c) The optimized grid while fixing the control points.}
	\label{fig:grid}
\end{figure}

Using the reshaped 3D faces as guidance, we warp each portrait image to generate the face with the target shape in 2D while avoiding visible artifacts. We first place a uniform grid $M_u=\{\bold{u}_i\}$ over the image as a warping proxy (see Figure~\ref{fig:grid} (a)), where $\bold{u}_i$ denotes the 2D coordinates of the grid point on the image. Then we propose a new method to find a set of control points $\{\bold{u}_c\} \in M_u $, which drive the image warping induced by the face deformation. Finally, we employ a least-square optimization to warp all the other grid points by minimizing the distortion of the entire grid.

\textbf{Control points selection.} It is straightforward to select control points directly on the 3D face model.
Jain et al. \shortcite{Jain2010} selected face mesh vertices and used their 2D projections as control points.
However, some control points (mesh vertices) may be occluded (or may emerge) after face deformation and projection, which can significantly affect the warped face shape on the image and cause severe artifacts.
Xiao et al. \shortcite{XiaoPortraits} utilized the closest grid points to the 2D face contour as control points. While their method works well for a single portrait, it ignores the occluded control points especially when enlarging a face.
Therefore, such a method cannot ensure that the control points are still at the face contour after reshaping, which leads to noticeable inconsistent and incoherent video frames after warping.


As the original video is already continuous with the right content, after face reshaping, the control points need to be stabilized at the location that expresses the same semantic information. 
Hence the problem of selecting control points amounts to find a mapping between face shape and control points before and after reshaping.

\textbf{SDF based selection.} We propose a SDF based method to establish consistent mapping of the image contour points before and after reshaping, and then transfer the mapping to control points. The benefit of stabilized face reconstruction is that the face contour in 3D is always complete, and its 2D projection gradually changes, which can be leveraged to achieve consistent and coherent control points selection. 
Note that unlike in 3D, semantically accurate face contour mapping in 2D does not always exist due to possible occlusions after reshaping and projection (e.g., face boundary partly blocked by nose but not as before), thus we employ an approximate mapping in practice as follows.

\begin{figure}[b]
	\centering

    \subfigure[Before reshaping]{
    \begin{minipage}[b]{.25\linewidth}
    \centering
    \includegraphics[scale=0.12]{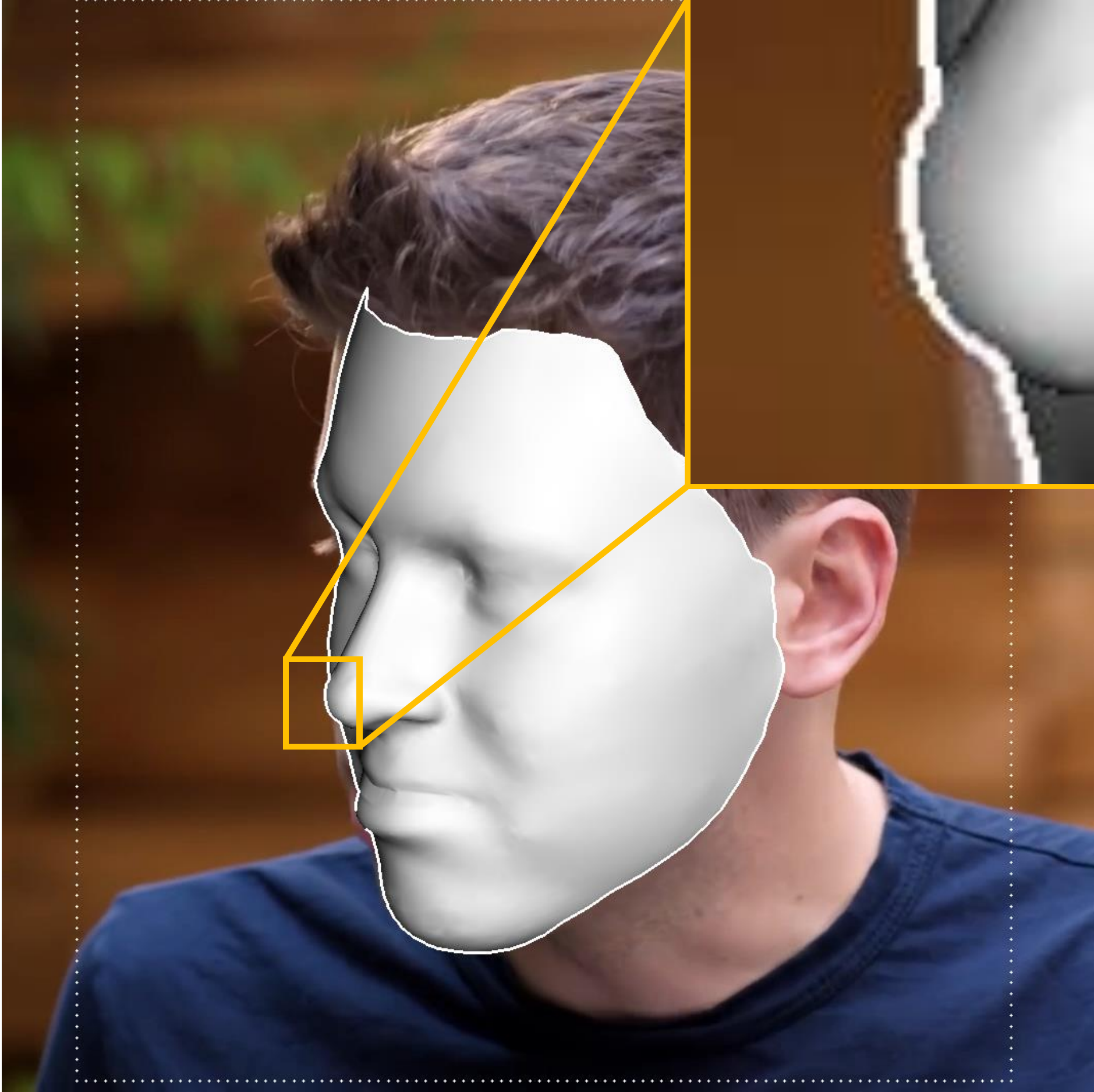}
    \end{minipage}\hspace{-3mm}
    }
    \subfigure[After reshaping]{
    \begin{minipage}[b]{.25\linewidth}
    \centering
    \includegraphics[scale=0.12]{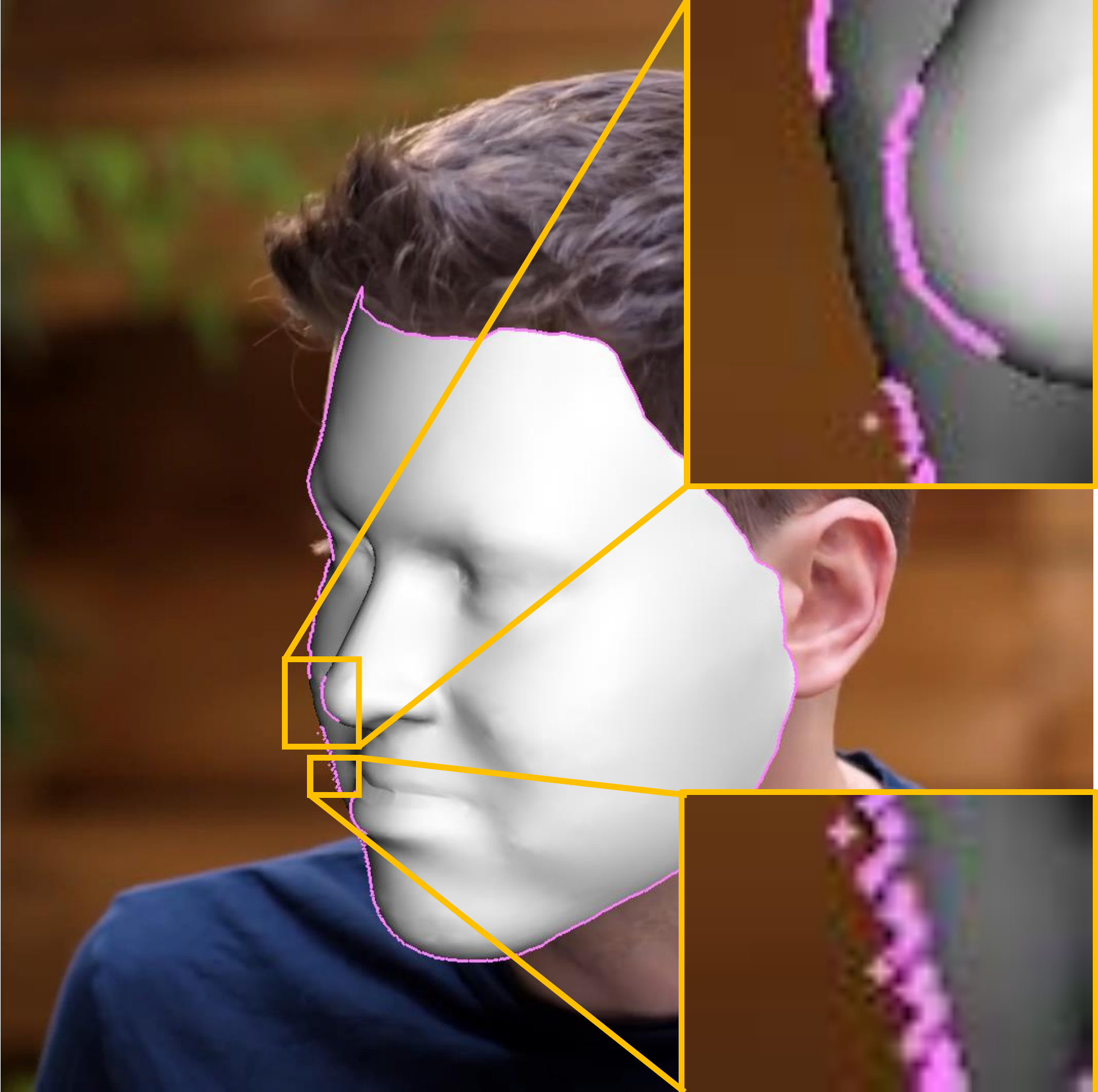}
    \end{minipage}\hspace{-3mm}
    }
    \subfigure[Without 3D information]{
    \begin{minipage}[b]{.25\linewidth}
    \centering
    \includegraphics[scale=0.12]{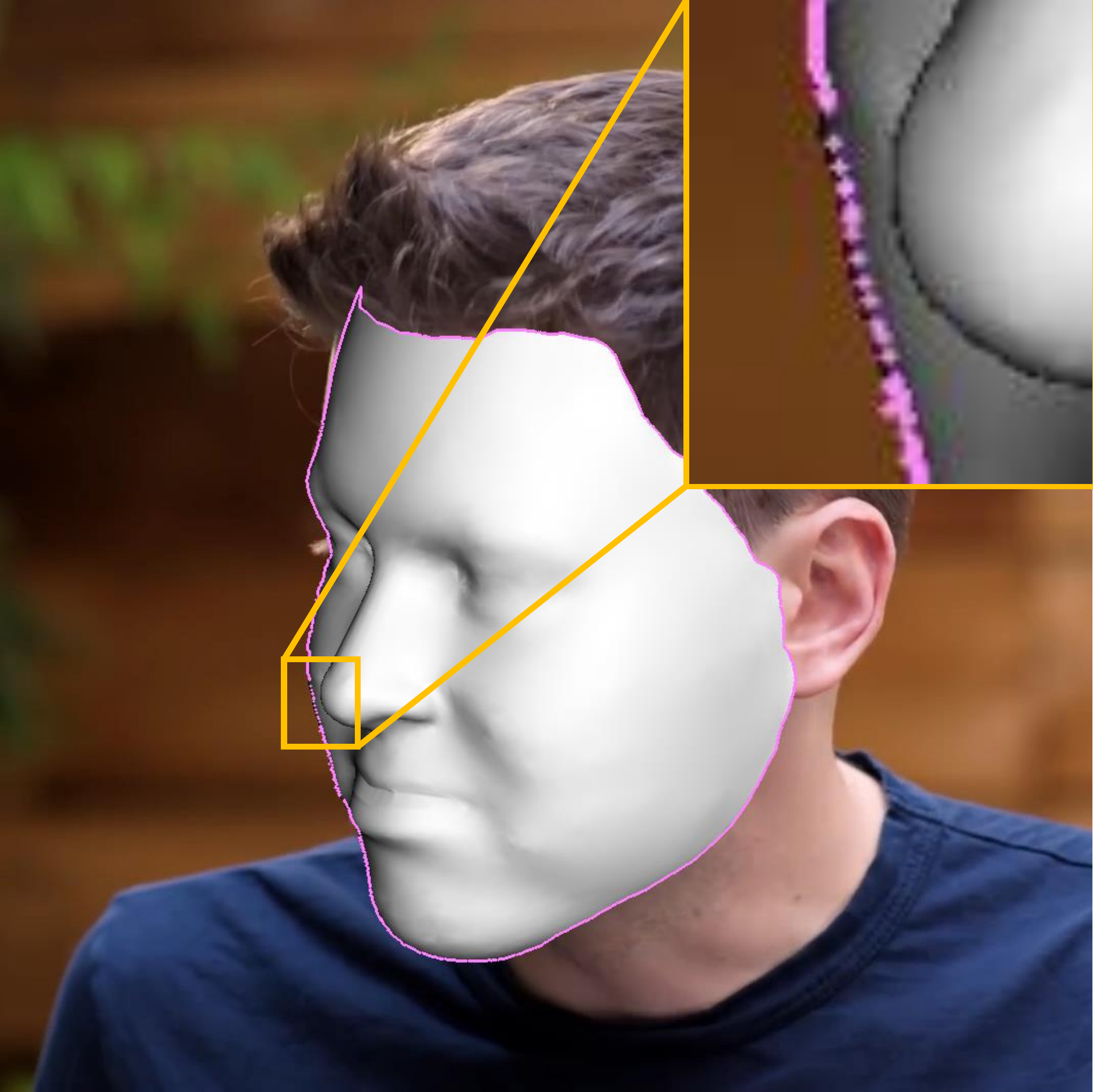}
    \end{minipage}\hspace{-3mm}
    }
    \subfigure[Our mapping]{
    \begin{minipage}[b]{.25\linewidth}
    \centering
    \includegraphics[scale=0.12]{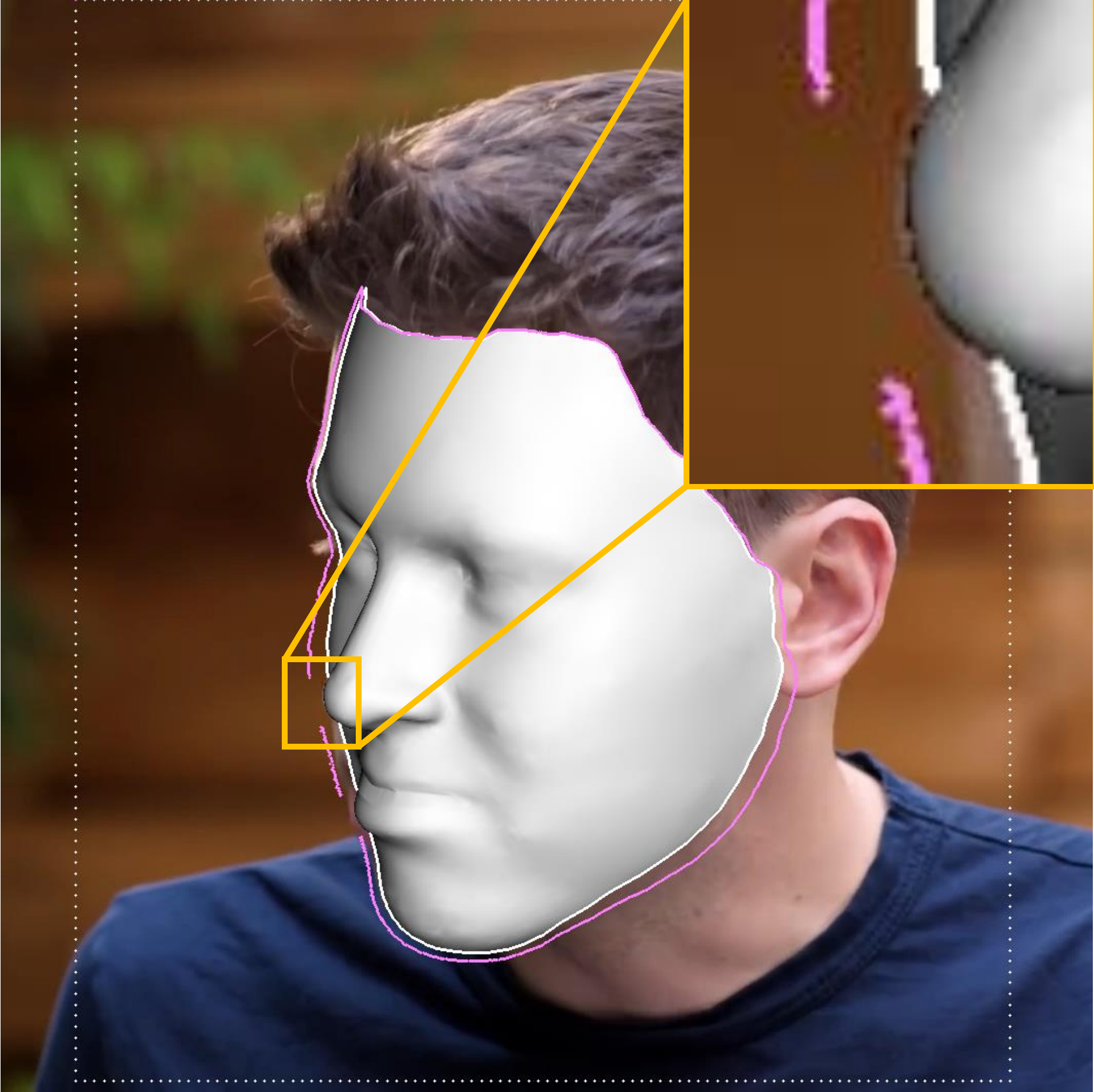}
    \end{minipage}
    }
    
	\caption{The white points in (a) are the unprojected 3D contour points before reshaping. The pink points in (b) are the 3D vertices corresponding to white points in (a) after reshaping, which causes inconsistent mapping. Figure (b) shows two kinds of inconsistency caused by occlusions after reshaping: 1) between nose and cheek, and 2) between cheeks. The pink points in (c) represent the corresponding contour points after reshaping without considering occlusion, which has established wrong mappings compared with (a). (d) is the result of our mapping method where the white contour is the same with (a) and the pink contour is the reshaped 3D model contour. Note that the wrong mapping points are successfully removed here.}
	\label{fig:nose}
\end{figure}

We first choose all face contour pixels of the original frame and map them to the reshaped face in 2D. We find that dense mapping of all contour pixels can further reduce inconsistency compared to sparse mapping of a subset of pixels (see comparison in Figure \ref{fig:contoursparse} and discussion in Section \ref{section:Evaluation}). We use the reconstructed 3D face model to guide the dense mapping. For each contour pixel, we first unproject it to the reconstructed 3D face model, then reshape the face model and project it back to the image to get its mapped pixel. This process cannot achieve consistent contour of 3D model after reshaping (see Figure~\ref{fig:nose} (b)). Therefore, we revise the incorrect mapping (by checking if the mapped pixel lies on the reshaped face contour) based on the SDF of the original 2D face contour (see pink contour in Figure~\ref{fig:nose} (c)). More specifically, for such a contour pixel, we move it along the gradient of the SDF until it meets the mapped pixel on the reshaped face contour in 2D (boundary of the projected 3D face after reshaping). This method still has problems in extreme cases where the mapping incorrectly maps nose points to cheeks points (see the boxed-out area both in Figure~\ref{fig:nose} (a) and (c)). For such cases, we remove the incorrect mappings according to the 3D structure information of each points (see Figure~\ref{fig:nose} (d)).


\textbf{Warping.} Based on the dense contour mapping, we select grid points $\{\bold{u}_c\}$ which are closest to the face contour. Then we employ moving least squares (MLS) deformation \cite{schaefer2006image} to find the target position of $\{\bold{u}_c\}$ guided by the dense mapping. After that, we employ a least-square optimization to other grid points while fixing the control points to minimize the overall distortion (see Figure~\ref{fig:grid}). We adds the linear bending term $E_l$ and regularization term $E_r$ proposed by Shih et al. \shortcite{shih2019distortion}:
\begin{equation}
E_l = \sum_i\sum_{j\in N(i)} ||(\bold{v}_i-\bold{v}_j) \times \bold{e}_{ij}||^2_2,
\end{equation}
\begin{equation}
E_r = \sum_i\sum_{j\in N(i)} ||\bold{v}_i-\bold{v}_j||^2_2,
\end{equation}
where $\bold{v_i}$ is the $i$-th key point of grid $M_u$, $N(i)$ is the set of 4-way adjacent points of key point $\bold{v}_i$, and $\bm{e}_{ij}$ is the unit vector along the direction $\bold{v}_i-\bold{v}_j$.

We only select the area surrounding the face as the target area for optimization. In order to make the optimized area blend perfectly with the non-optimized area, we fix the boundary grid points instead of adding a grid border term as in \cite{shih2019distortion} to adjust the distortion in the boundary. The final energy function is:
\begin{equation}
E=w_lE_l+w_rE_r.
\label{eq:optimization}
\end{equation}
We empirically set the weights as $w_l=1$, $w_r=0.8$.


\section{Evaluation}\label{section:Evaluation}

In this section, we extensively evaluate the proposed method, including testing on various examples, comparing with image reshaping based method, validating our method design choices with baselines, and presenting the performance and
implementation details.

\subsection{Results}
We test our method on a variety of portrait videos with different  gender, hairstyle, skin color, etc. Figure~\ref{fig:teaser} shows some sampled video frames generated by our method. It can be seen that our method can successfully reshape  portraits without introducing visual artifacts. By changing the parameter $\delta$, we can continuously adjust the portrait shape (either thinner or rounder) for potential face retouching and exaggeration applications. More examples and full video sequences can be found in the supplemental video.

\begin{figure}[b]
	\centering
	\subfigure[Initial frame]{
	\includegraphics[width=0.3\linewidth]{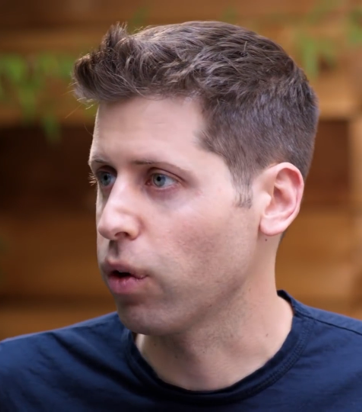}
	\label{fig:InitialCommpare}
	}%
    \subfigure[Xiao et al. \shortcite{XiaoPortraits}]{
        \includegraphics[width=0.3\linewidth]{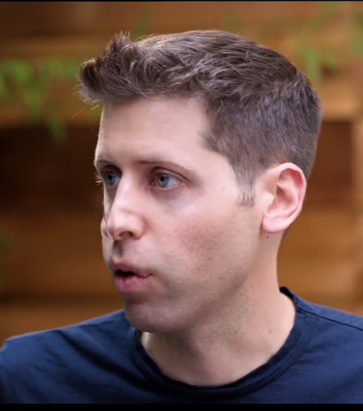}
        \label{fig:XiaoMethod}
    }%
    \subfigure[Our method]{
        \includegraphics[width=0.3\linewidth]{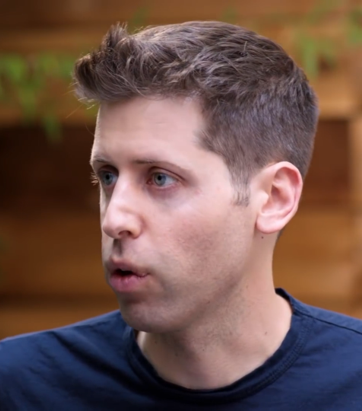}
        \label{fig:OurCompare}
    }%

	\caption{Comparison with the baseline method. Given a frame from a portrait video (a), Xiao et al. \shortcite{XiaoPortraits}'s  reshaping method produces artifacts (b) near the tip of the nose because the nose occludes the side face, while our method can still generate satisfactory results (c) with the same reshaping parameter.}
	\label{fig:compareXiao}
\end{figure}

\subsection{Comparison}

As there is no prior work on parametric reshaping of portraits in videos, we take the parametric reshaping method of portrait images \cite{XiaoPortraits} as the baseline for comparison. Given a portrait video footage, for a frame shown in Figure \ref{fig:InitialCommpare}, as the nose occludes the side face of the portrait, the method of Xiao et al. \shortcite{XiaoPortraits} will produce obvious artifacts (see Figure \ref{fig:XiaoMethod}) near the tip of the nose because of their sparse mapping, which is successfully solved using our dense mapping method (see Figure \ref{fig:OurCompare}) with the same reshaping parameter. Moreover, their method cannot generate smooth and coherent reshaping portrait video results. Please refer to the accompanying video for side-by-side comparison. The results show that our approach can robustly produce coherent reshaped portrait videos while the image-based method can easily lead to noticeable flickering artifacts.

\begin{figure}[b]
	\centering
	\includegraphics[width=1\linewidth]{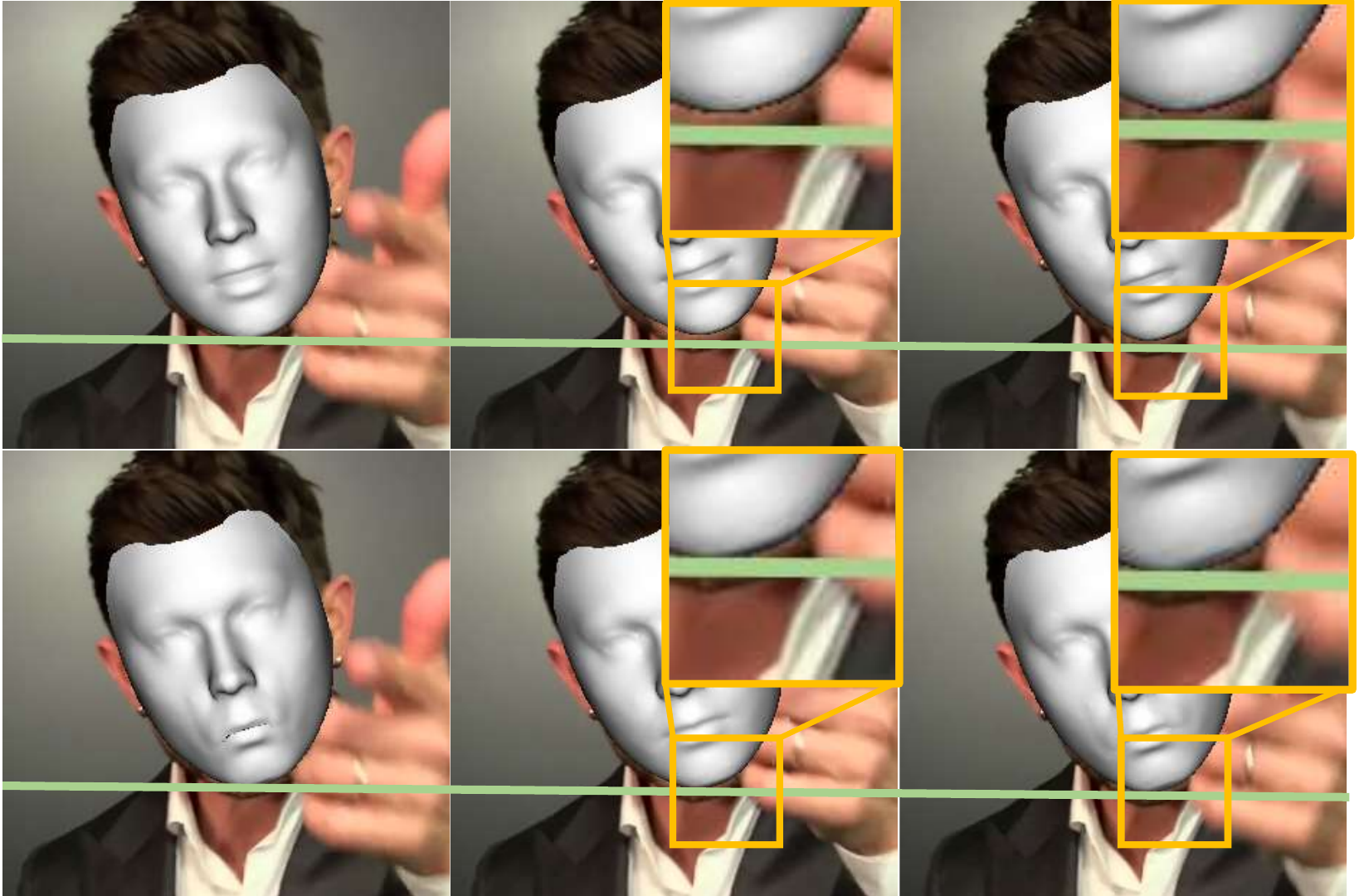}
    \caption{ The results of using the naive dense flow energy defined on face contours (top row). Results by using our dense flow energy (bottom row). The green line connects the chin bottom of the first and third images. By referring to the green line, we can see that the face models in the second-row have jittering effects (as shown by the gap between the green line and the chin bottom).}
	\label{fig:est_trans}
\end{figure}

\subsection{Ablation studies}
We also verify the design choices of our method by validating its two main stages, i.e., face reconstruction and video reshaping, respectively. We show the effectiveness of our design choices by comparing them with other baselines.

%

\begin{figure}[t]
	\centering
	\subfigure[Input portrait image]{
	\includegraphics[width=0.33\linewidth]{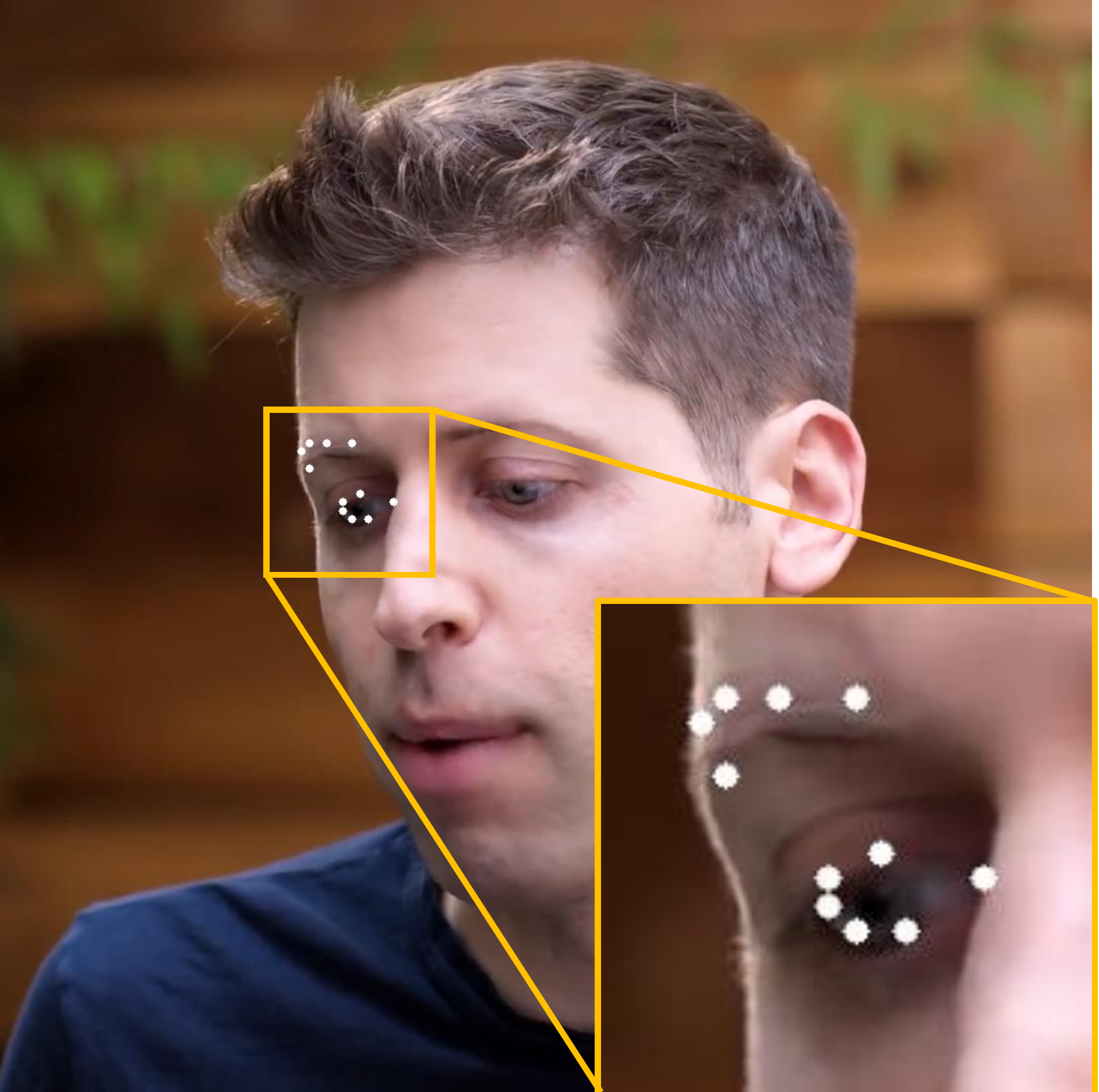}
	\label{fig:initial_align}
	}%
    \subfigure[Without constraint]{
        \includegraphics[width=0.33\linewidth]{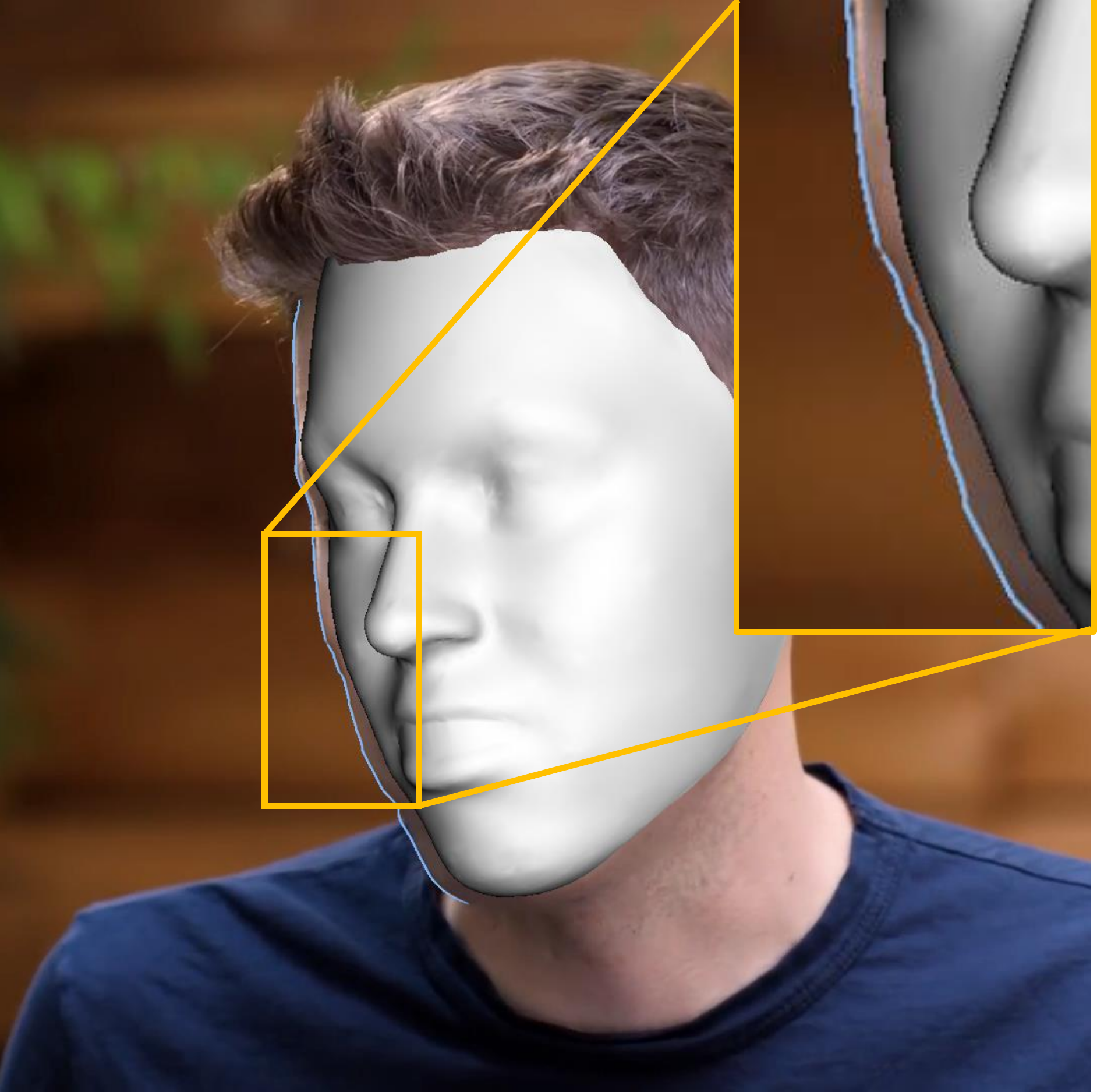}
        \label{fig:no_align}
    }%
    \subfigure[Our method]{
        \includegraphics[width=0.33\linewidth]{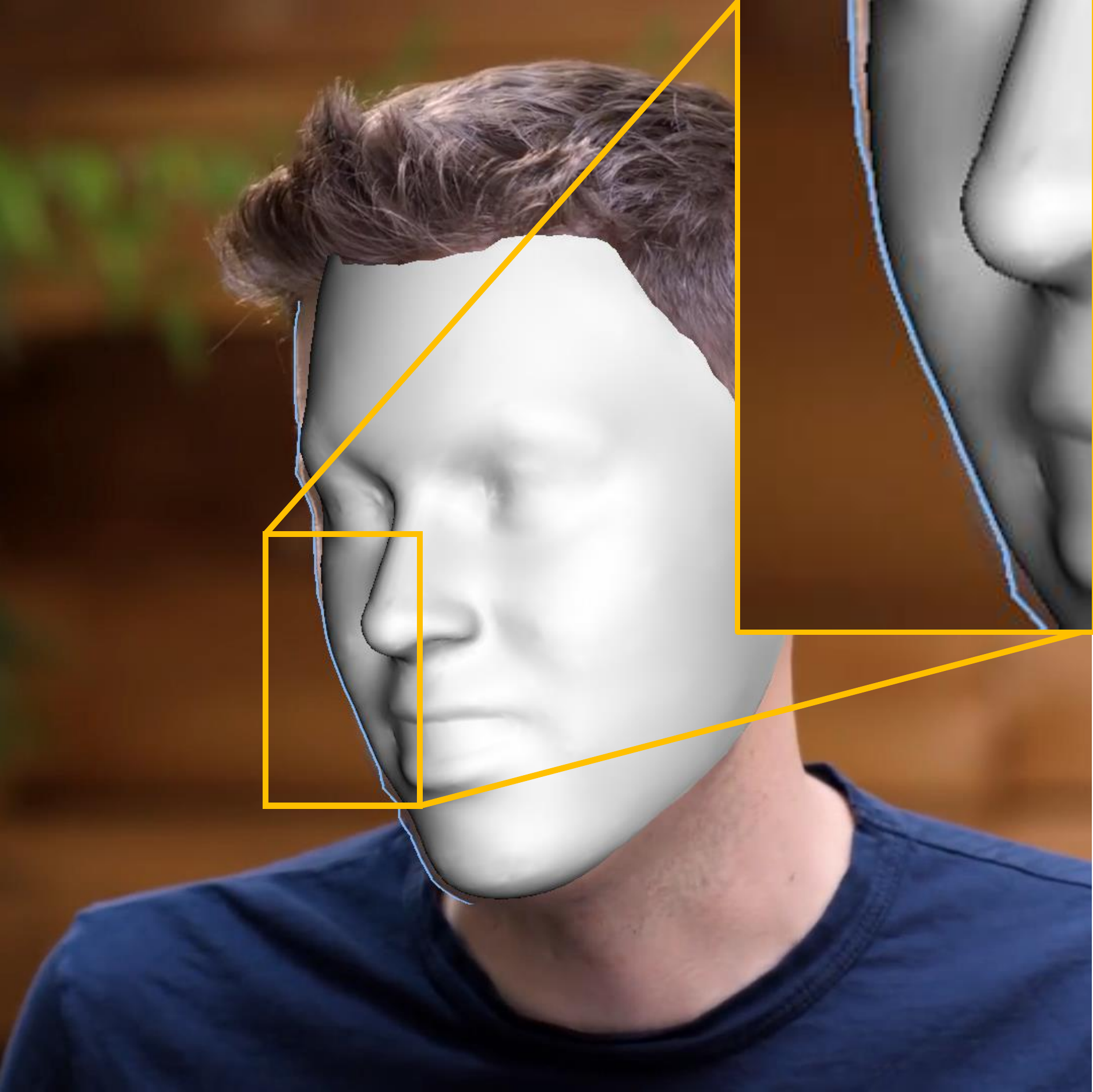}
        \label{fig:our_align}
    }%

	\caption{Comparison between the method without constraining face contours (b) and ours (c). White points in (a) are detected as landmarks having offset to the left near the eye. The method without contour constraints results in a worse alignment of face contour (see gap in-between 2D and 3D).}
	\label{fig:contour}
\end{figure}

\subsubsection{Face reconstruction} 


We first evaluate the effectiveness of our new dense flow energy. We compare our result to the result without using dense flow energy, and the result using dense flow energy but defined on the entire face. 

In most cases, using dense flow energy helps to produce visually pleasing results. The stability difference between using dense flow energy or not is better to be seen in video sequence. It is worth noting that the dense flow energy can effectively suppress the effects of jittering landmarks near the face contour, but cannot eliminate them completely. If the landmarks themselves are jittering, the estimated transformations cannot be very smooth.
Figure \ref{fig:est_trans} shows the problematic case without discarding vertices inside the wrong regions of the flow map. The handshakes near the face cause wrong motion directions in the flow map, resulting in instability head transformations.

%
We then evaluate the contour energy which restricts the face model to align with the 2D face contour. It is more difficult to detect the landmarks of the side face compared to the frontal one. We found that the detected landmarks tend to have offset at the side when they are far away from the camera, leading to a poor alignment between the face model and its 2D face contour. For example, the landmarks near the eye have offset as shown in Figure \ref{fig:initial_align}. Figure \ref{fig:no_align} and Figure \ref{fig:our_align} show results without and with the contour energy, respectively.  Although both results have similar eye locations, the one using contour energy achieves better alignment. For an image sequence, aligned contours result in continuous and stable face transformations.

\subsubsection{Video reshaping}
\begin{figure}[b]
	\centering
	\subfigure[MLS-only]{
	\includegraphics[width=0.245\linewidth]{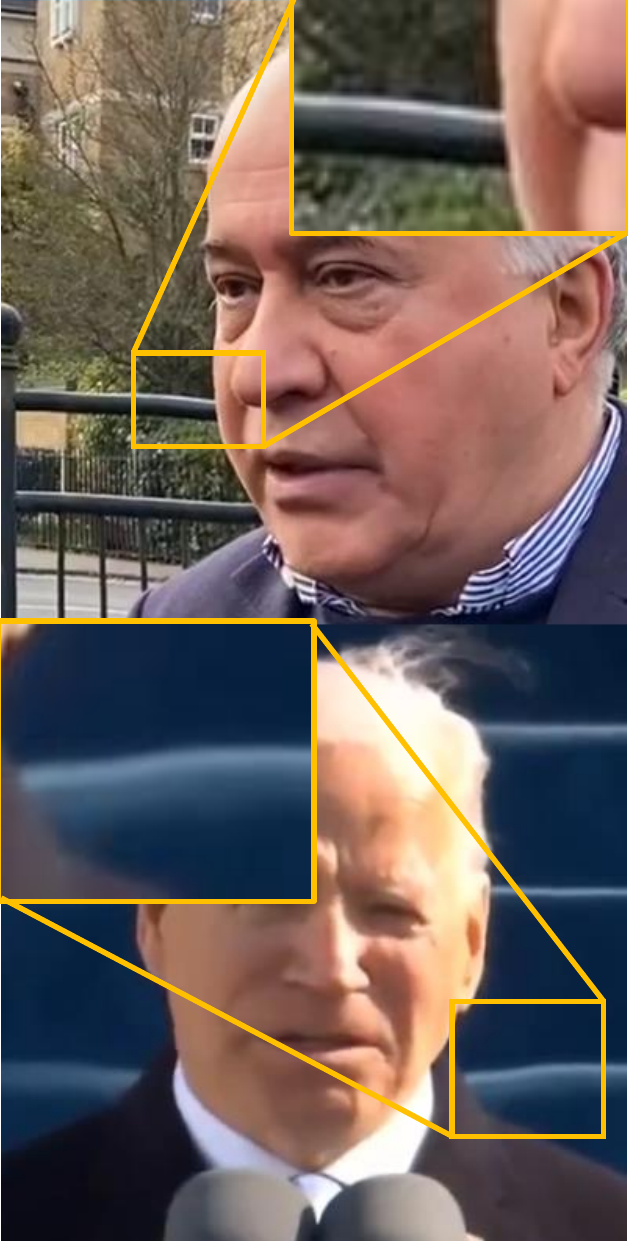}
	}%
    \subfigure[Optimization-only]{
        \includegraphics[width=0.245\linewidth]{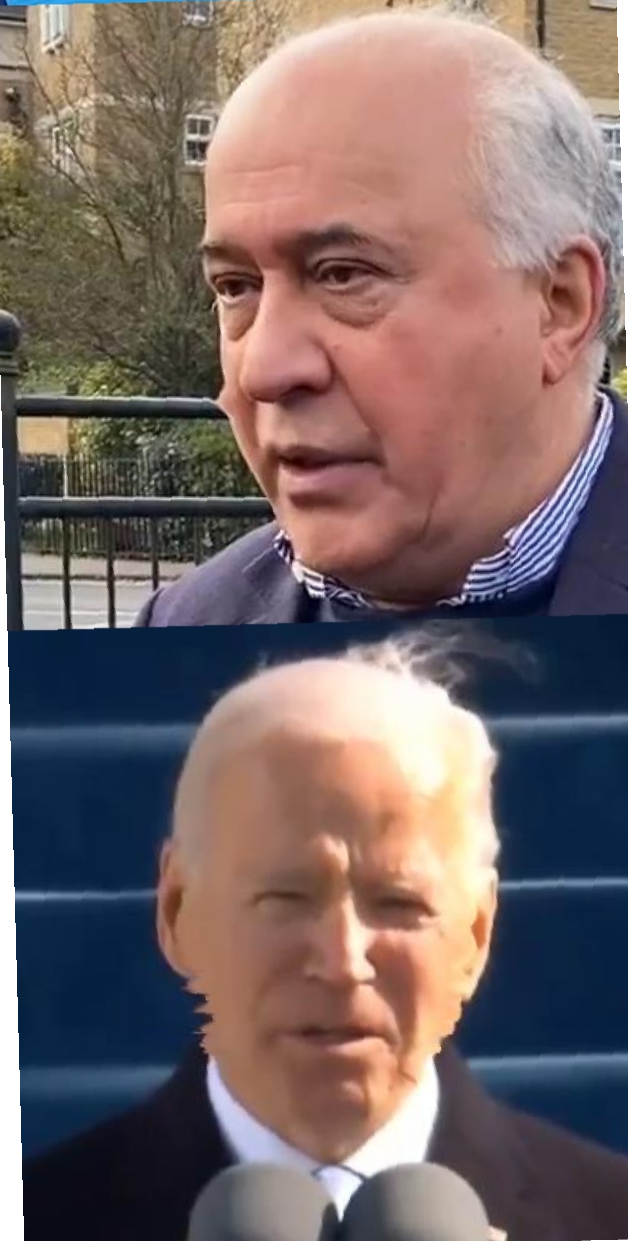}
    }%
    \subfigure[Our method]{
        \includegraphics[width=0.487\linewidth]{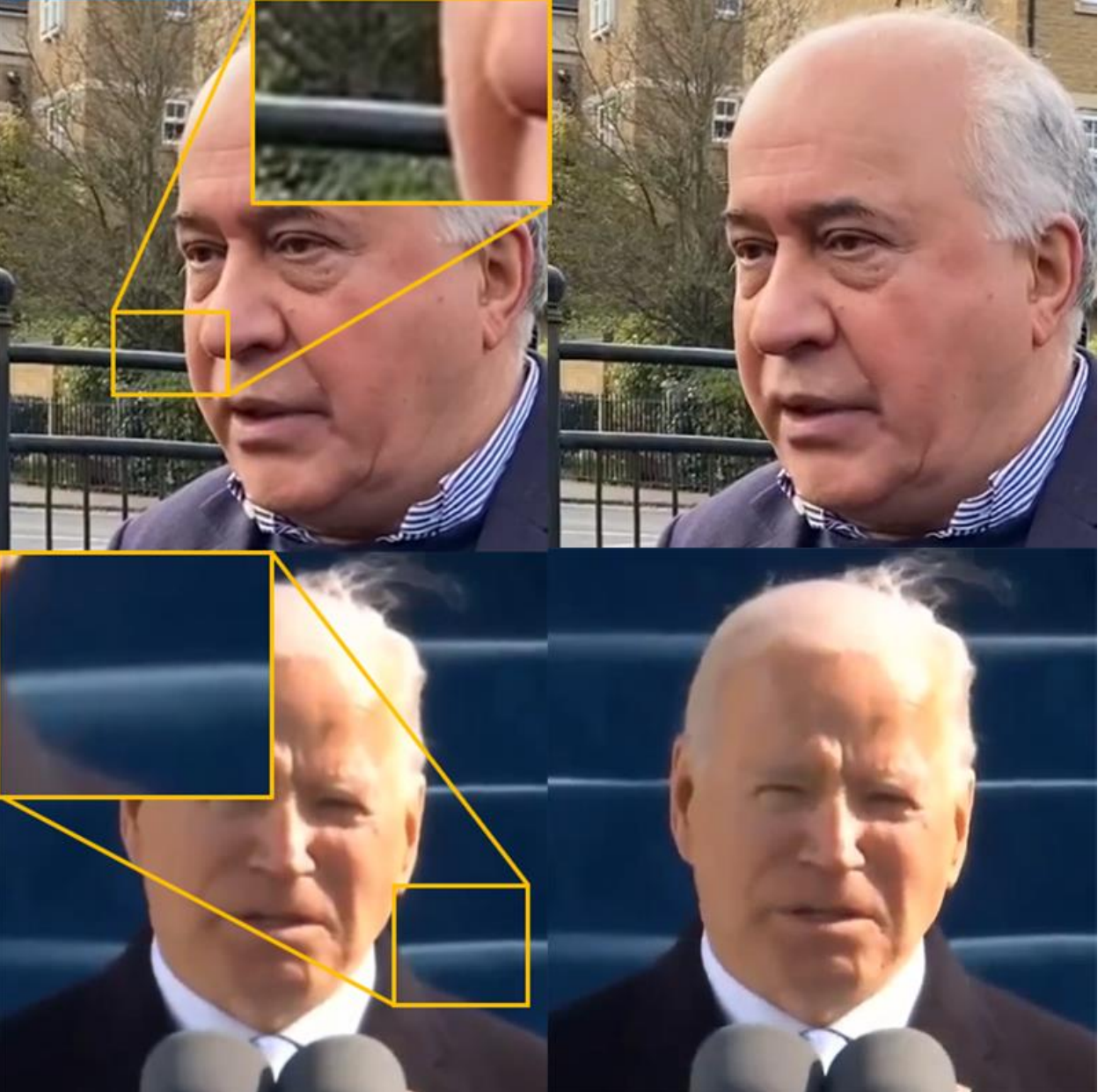}
    }%

	\caption{Comparison between the method using MLS only (a), the method using optimization only (b), and ours (c). Our method achieves better results in terms of background distortion and face boundary coherence.}
	\label{fig:mlsoptimize}
\end{figure}

We first perform comparisons to demonstrate that our approach which combines MLS and grid optimization is meaningful and effective. Figure \ref{fig:mlsoptimize} shows the comparison with the MLS-only approach, the optimization-only approach, and our approach. It shows that the grid optimization is effective in correcting background distortion. 
And the MLS approach ensures the coherence of the face boundaries and the video stability. Besides, optimization-only approach requires a higher grid resolution (up to four times) to achieve similar results, but still has incoherent face boundary.

\begin{figure}[t]
\setlength{\abovecaptionskip}{0.cm}
\setlength{\belowcaptionskip}{0.cm}
\setlength{\belowdisplayskip}{0.cm}
\vspace{-0.8cm}
	\centering
	\subfigure[Without fixing contour]{
	\includegraphics[width=0.33\linewidth]{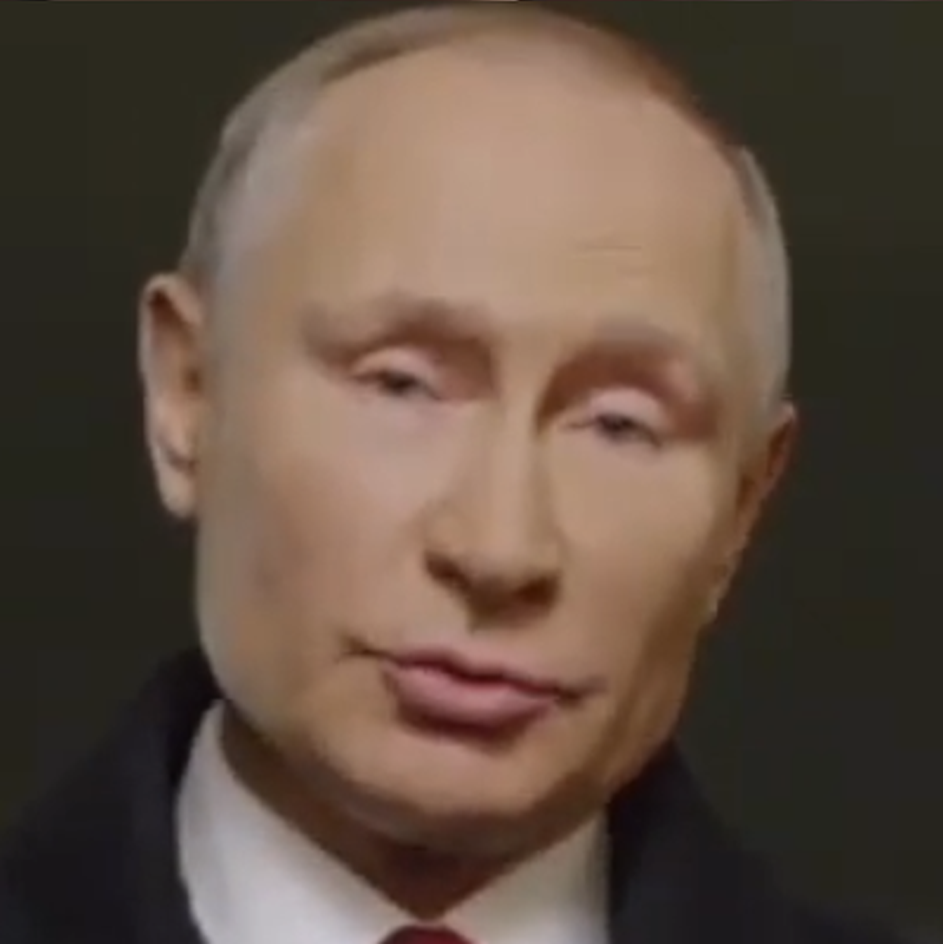}
	}%
    \subfigure[Sparse mapping]{
        \includegraphics[width=0.33\linewidth]{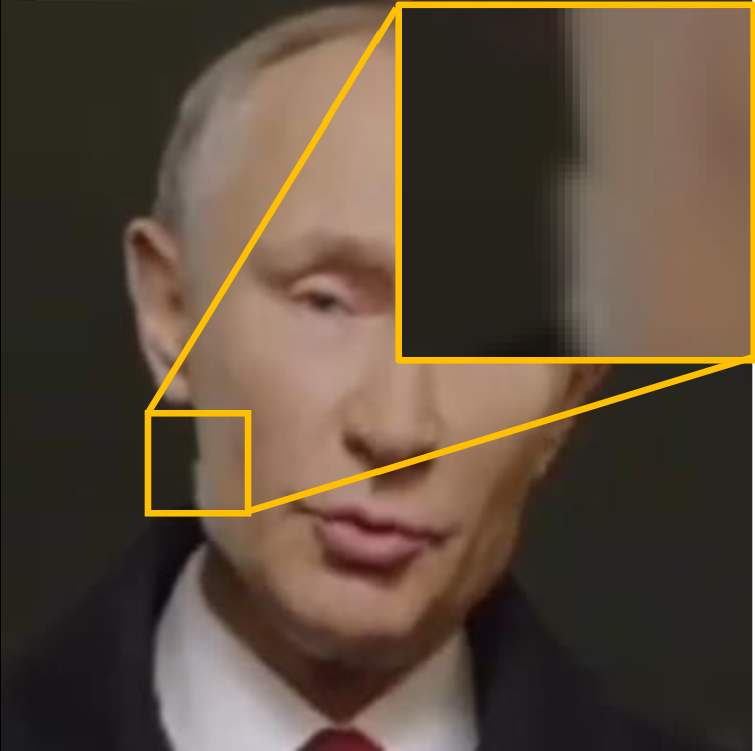}
    }%
    \subfigure[Our method]{
        \includegraphics[width=0.33\linewidth]{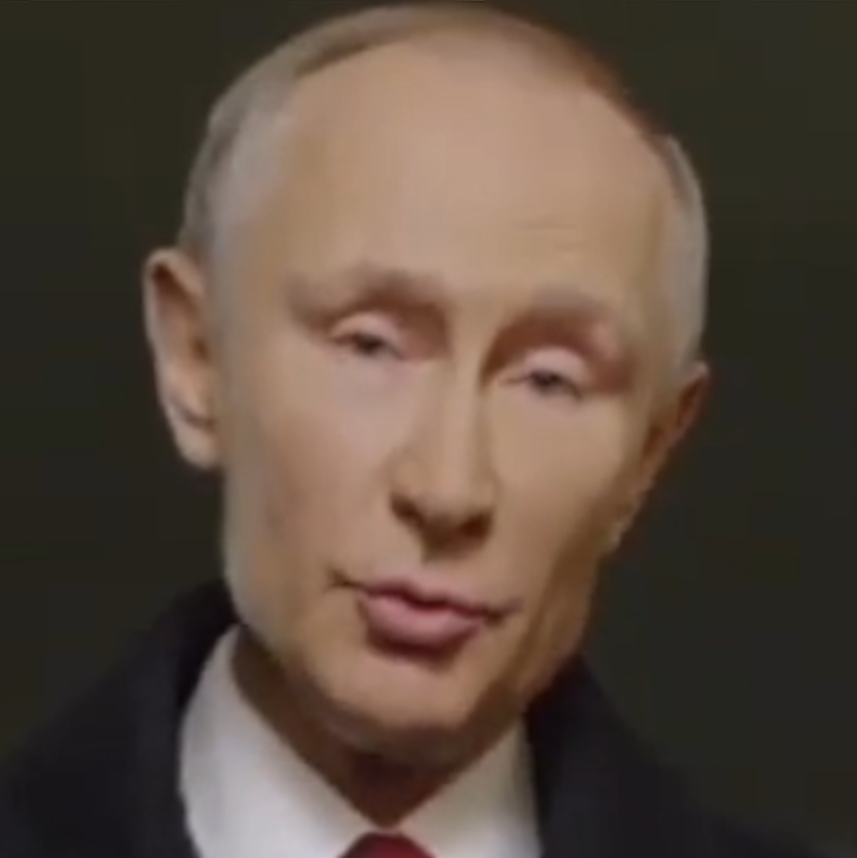}
    }%
	\caption{Comparison of extreme reshaping results between the method that does not fix contour grid points (a), the method that only uses sparse contour points mapping (b), and ours (c). Our method achieves better results in terms of face boundary coherence and reshaping consistency. 
	}
	\label{fig:contoursparse}
	
\end{figure}

We then perform comparisons to prove that using SDF based dense boundary mapping is effective for preserving face reshaping consistency and the stability of the video.  Figure \ref{fig:contoursparse}(a) shows the case of not fixing face contour points, where only the feature points used for reshaping are exploited as fixed control points. This method does 
not reflect enough information about the reshaping, thus the obtained results do not coincide with the projection of the 3D model after reshaping. Figure \ref{fig:contoursparse}(b) shows the case of only using a set of sparse grid points to represent the face boundary. Here we select suitable initial grid points coincided with the face contour as fixed control points. {Two more energy terms are introduced in Eqn.~{\ref{eq:optimization}} in order to obtain reasonable results (see Appendix for details).}

The results are better than other baselines. In the case of a small degree of face reshaping, the defects are almost invisible and it can generally achieve video coherence. However, if the face is largely deformed, as shown in Figure \ref{fig:contoursparse}(b), protrusions and gaps in the face boundary appear and they may affect video continuity. Also, the time consumed by this method is significantly increased.

\begin{figure}[t]
	\centering
    \setlength{\abovedisplayskip}{1pt}
    \setlength{\belowdisplayskip}{1pt}
	\subfigure[Without using 3D information]{
	\includegraphics[width=0.5\linewidth]{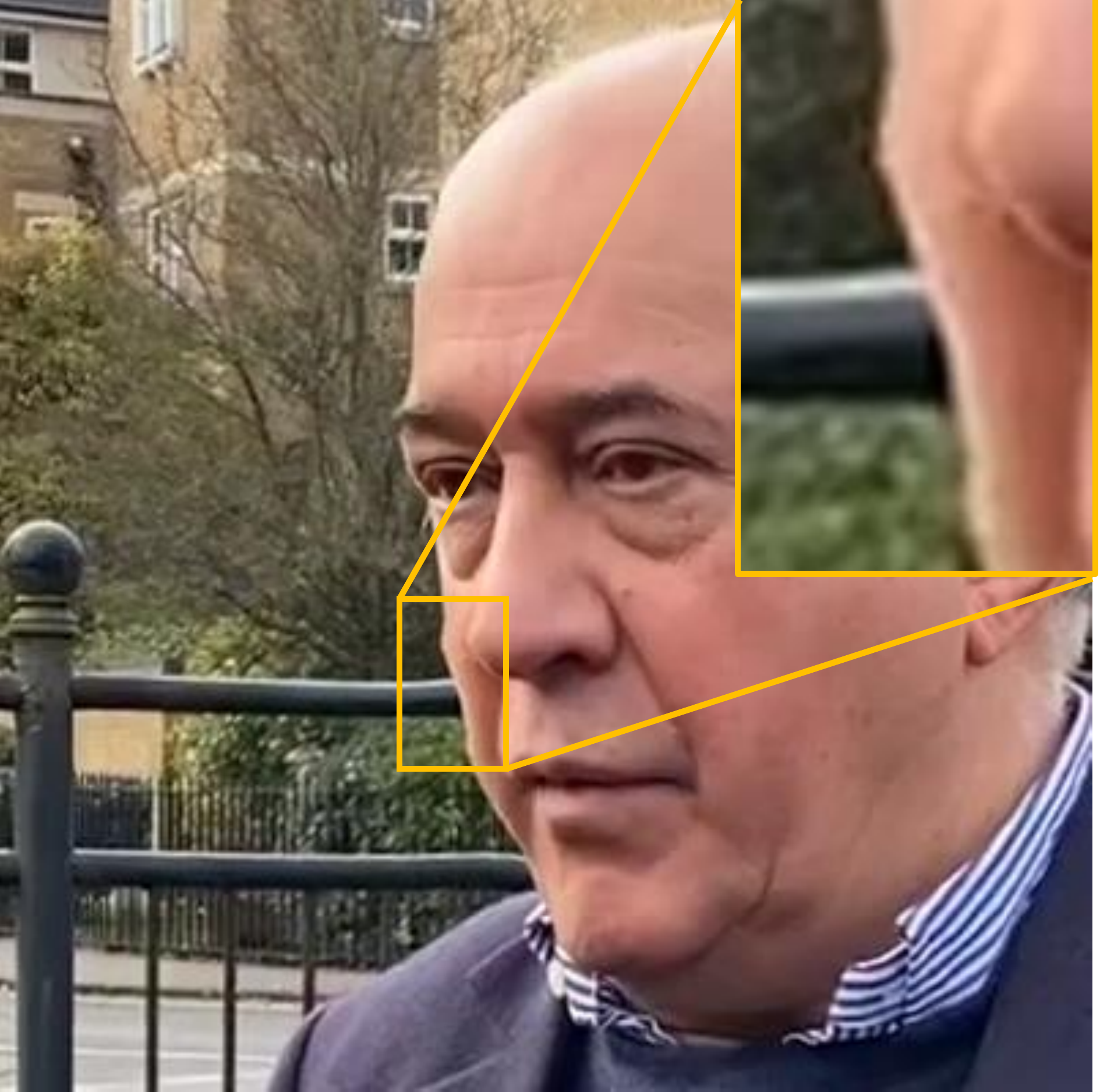}
	}%
    \subfigure[Our method]{
        \includegraphics[width=0.5\linewidth]{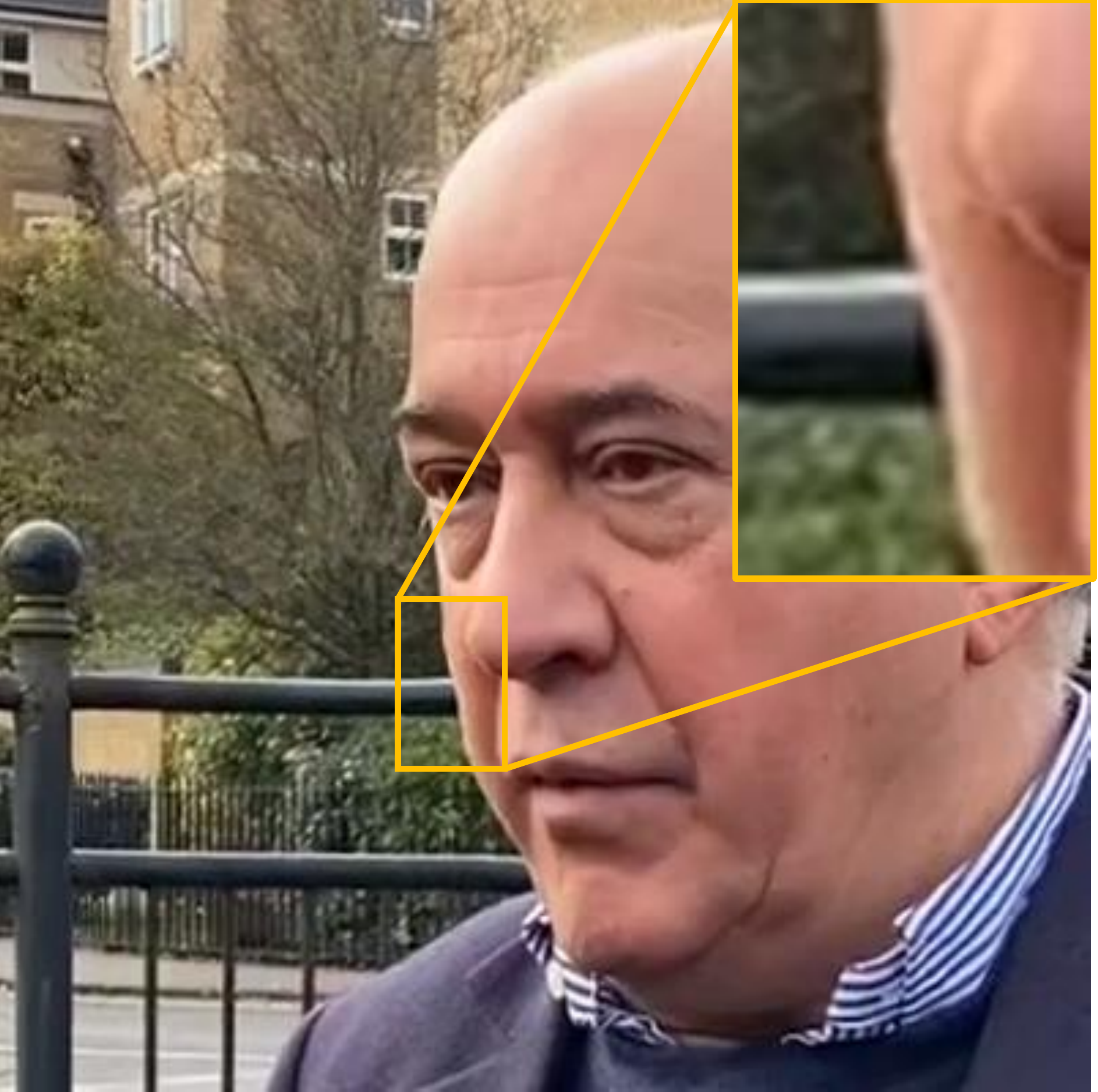}
    }%
	\caption{Comparison of extreme reshaping results between the method that does not use 3D information to establish the dense mapping (a) and ours (b). 	
	Our method performs better at face boundary.	
	}
	\label{fig:findboundary}
\end{figure}

Figure \ref{fig:findboundary} shows the result of selecting control points based on dense mapping from SDF without using the 3D model information. This performs well in most cases, but has problem when the nose contour affects the face contour.

\subsection{Implementation Details}\label{section_Implementation}
 We use the optical flow algorithm provided by OpenCV to extract the motion map, and use Gaussian sampling to get the motion value $\bold{U}_i$ of each projected face vertex instead of bilinear interpolation to avoid local minima and obtain smoother results~\cite{ren2019structureflow}.
 We perform facial landmark detection as proposed in \cite{bulat2017far}.
We use Ceres \cite{ceres-solver} to solve optimization in Section \ref{section:reconstruction} and Section \ref{section:reshaping}. For image warping, we set the grid dimension to $100\times 100$ for all of our results.

Our implementation is on a desktop PC with AMD Ryzen 9 3950X CPU and 32GB memory. In the face reconstruction stage, pose estimation takes 120ms per frame on average, identity estimation computes only once and takes 321ms, and the expression estimation takes 150ms per frame. In the video reshaping stage, the 3D face deformation takes 160ms at the beginning. Image warping takes 266ms per frame, where MLS based image deformation and grid optimization take 71ms and 74ms, respectively.

The videos we used are all downloaded from public datasets or Youtube websites. The sources of each video are listed in Appendix.

\section{Discussions}


All of our results show plausible reshaping without visible artifacts. However, our method still has some limitations. First, visual distortion may appear in the surrounding regions close to the face when large shape deformation is employed. Video inpainting methods \cite{kim2019deep} could be employed on background regions to further improve the results. Second, we note the fact that wrinkles will reduce and a double chin will appear when gaining weight on the face; while wrinkles will increase and a double chin will diminish when losing weight. But similar to reshaping portrait images \cite{XiaoPortraits}, our approach cannot deal with such changes. 
Further, our current approach is fully unsupervised, which means it does not require any customized face priors, thus it can be directly used to process portrait videos in the wild.
On the other hand, with the help of more pre-knowledge such as face shape, our method can be consequently adapted and accelerated to reshape portrait videos on the fly.


%

\section{Conclusions}

We have presented the first method to reshape portraits in a video. Our video-based face reconstruction method is able to eliminate the effects of jittering landmarks and incorrect flow map, resulting in a steady 3D face model sequence with accurate identities and smooth transformations. We achieve a consistent video deformation by aligning the face model sequence with the face contours of the image sequence. We employ an SDF based approach to produce a dense and smooth mapping from the initial face to the reshaped face, which effectively minimizes the warping distortion and avoids visual artifacts after video reshaping. Extensive evaluations and comparisons demonstrate that our method can generate high-quality portrait video reshaping results.

\begin{acks}
Xiaogang Jin was supported by  the National Natural Science Foundation of China (Grant No. 61972344), the Key Research and Development Program of Zhejiang Province (Grant no. 2018C03055), and the Ningbo Major Special Projects of the “Science and Technology Innovation 2025” (Grant No. 2020Z007). Yong-Liang Yang was partly supported by
RCUK grant CAMERA (EP/M023281/1, EP/T022523/1), and a gift
from Adobe.
\end{acks}


\bibliographystyle{ACM-Reference-Format}
\balance
\normalem\bibliography{sample-base}


\begin{thebibliography}{30}


\ifx \showCODEN    \undefined \def \showCODEN     #1{\unskip}     \fi
\ifx \showDOI      \undefined \def \showDOI       #1{#1}\fi
\ifx \showISBNx    \undefined \def \showISBNx     #1{\unskip}     \fi
\ifx \showISBNxiii \undefined \def \showISBNxiii  #1{\unskip}     \fi
\ifx \showISSN     \undefined \def \showISSN      #1{\unskip}     \fi
\ifx \showLCCN     \undefined \def \showLCCN      #1{\unskip}     \fi
\ifx \shownote     \undefined \def \shownote      #1{#1}          \fi
\ifx \showarticletitle \undefined \def \showarticletitle #1{#1}   \fi
\ifx \showURL      \undefined \def \showURL       {\relax}        \fi
\providecommand\bibfield[2]{#2}
\providecommand\bibinfo[2]{#2}
\providecommand\natexlab[1]{#1}
\providecommand\showeprint[2][]{arXiv:#2}

\bibitem[\protect\citeauthoryear{Agarwal, Mierle, and Others}{Agarwal
  et~al\mbox{.}}{2012}]%
        {ceres-solver}
\bibfield{author}{\bibinfo{person}{Sameer Agarwal}, \bibinfo{person}{Keir
  Mierle}, {and} \bibinfo{person}{Others}.} \bibinfo{year}{2012}\natexlab{}.
\newblock \bibinfo{title}{Ceres Solver}.
\newblock \bibinfo{howpublished}{\url{http://ceres-solver.org}}.
\newblock


\bibitem[\protect\citeauthoryear{Blanz and Vetter}{Blanz and Vetter}{1999}]%
        {BlandVetter99}
\bibfield{author}{\bibinfo{person}{Volker Blanz} {and} \bibinfo{person}{Thomas
  Vetter}.} \bibinfo{year}{1999}\natexlab{}.
\newblock \showarticletitle{A morphable model for the synthesis of 3D faces}.
  In \bibinfo{booktitle}{\emph{the 26th Annual Conference on Computer Graphics
  and Interactive Techniques}}. \bibinfo{publisher}{{ACM}},
  \bibinfo{pages}{187--194}.
\newblock


\bibitem[\protect\citeauthoryear{Booth, Antonakos, Ploumpis, Trigeorgis,
  Panagakis, and Zafeiriou}{Booth et~al\mbox{.}}{2017}]%
        {Booth}
\bibfield{author}{\bibinfo{person}{James Booth}, \bibinfo{person}{Epameinondas
  Antonakos}, \bibinfo{person}{Stylianos Ploumpis}, \bibinfo{person}{George
  Trigeorgis}, \bibinfo{person}{Yannis Panagakis}, {and}
  \bibinfo{person}{Stefanos Zafeiriou}.} \bibinfo{year}{2017}\natexlab{}.
\newblock \showarticletitle{3D Face Morphable Models "In-the-Wild"}. In
  \bibinfo{booktitle}{\emph{{IEEE} Conference on Computer Vision and Pattern
  Recognition (CVPR)}}. \bibinfo{pages}{5464--5473}.
\newblock


\bibitem[\protect\citeauthoryear{Bulat and Tzimiropoulos}{Bulat and
  Tzimiropoulos}{2017}]%
        {bulat2017far}
\bibfield{author}{\bibinfo{person}{Adrian Bulat} {and}
  \bibinfo{person}{Georgios Tzimiropoulos}.} \bibinfo{year}{2017}\natexlab{}.
\newblock \showarticletitle{How Far are We from Solving the 2D {\&} 3D Face
  Alignment Problem? (and a Dataset of 230, 000 3D Facial Landmarks)}. In
  \bibinfo{booktitle}{\emph{{IEEE} International Conference on Computer Vision
  (ICCV)}}. \bibinfo{pages}{1021--1030}.
\newblock


\bibitem[\protect\citeauthoryear{Cao, Chai, Woodford, and Luo}{Cao
  et~al\mbox{.}}{2018}]%
        {Cao2018}
\bibfield{author}{\bibinfo{person}{Chen Cao}, \bibinfo{person}{Menglei Chai},
  \bibinfo{person}{Oliver~J. Woodford}, {and} \bibinfo{person}{Linjie Luo}.}
  \bibinfo{year}{2018}\natexlab{}.
\newblock \showarticletitle{{Stabilized Real-time Face Tracking via a Learned
  Dynamic Rigidity Prior}}.
\newblock \bibinfo{journal}{\emph{ACM Transactions on Graphics (TOG)}}
  \bibinfo{volume}{37}, \bibinfo{number}{6} (\bibinfo{year}{2018}),
  \bibinfo{pages}{1--11}.
\newblock


\bibitem[\protect\citeauthoryear{Chen, Chen, Ni, and Ge}{Chen
  et~al\mbox{.}}{2020}]%
        {ChenSimSwap}
\bibfield{author}{\bibinfo{person}{Renwang Chen}, \bibinfo{person}{Xuanhong
  Chen}, \bibinfo{person}{Bingbing Ni}, {and} \bibinfo{person}{Yanhao Ge}.}
  \bibinfo{year}{2020}\natexlab{}.
\newblock \showarticletitle{SimSwap: An Efficient Framework For High Fidelity
  Face Swapping}. In \bibinfo{booktitle}{\emph{Proceedings of the 28th ACM
  International Conference on Multimedia}}. \bibinfo{pages}{2003–2011}.
\newblock


\bibitem[\protect\citeauthoryear{Chen, Zhu, Shamir, and Hu}{Chen
  et~al\mbox{.}}{2013}]%
        {Chen2013}
\bibfield{author}{\bibinfo{person}{Tao Chen}, \bibinfo{person}{Jun{-}Yan Zhu},
  \bibinfo{person}{Ariel Shamir}, {and} \bibinfo{person}{Shi{-}Min Hu}.}
  \bibinfo{year}{2013}\natexlab{}.
\newblock \showarticletitle{Motion-Aware Gradient Domain Video Composition}.
\newblock \bibinfo{journal}{\emph{IEEE Transactions on Image Processing}}
  \bibinfo{volume}{22}, \bibinfo{number}{7} (\bibinfo{year}{2013}),
  \bibinfo{pages}{2532--2544}.
\newblock


\bibitem[\protect\citeauthoryear{Egger, Smith, Tewari, Wuhrer, Zollh{\"{o}}fer,
  Beeler, Bernard, Bolkart, Kortylewski, Romdhani, Theobalt, Blanz, and
  Vetter}{Egger et~al\mbox{.}}{2020}]%
        {Egger2020}
\bibfield{author}{\bibinfo{person}{Bernhard Egger}, \bibinfo{person}{William
  A.~P. Smith}, \bibinfo{person}{Ayush Tewari}, \bibinfo{person}{Stefanie
  Wuhrer}, \bibinfo{person}{Michael Zollh{\"{o}}fer}, \bibinfo{person}{Thabo
  Beeler}, \bibinfo{person}{Florian Bernard}, \bibinfo{person}{Timo Bolkart},
  \bibinfo{person}{Adam Kortylewski}, \bibinfo{person}{Sami Romdhani},
  \bibinfo{person}{Christian Theobalt}, \bibinfo{person}{Volker Blanz}, {and}
  \bibinfo{person}{Thomas Vetter}.} \bibinfo{year}{2020}\natexlab{}.
\newblock \showarticletitle{3D Morphable Face Models - Past, Present, and
  Future}.
\newblock \bibinfo{journal}{\emph{ACM Transactions on Graphics (TOG)}}
  \bibinfo{volume}{39}, \bibinfo{number}{5} (\bibinfo{year}{2020}),
  \bibinfo{pages}{157:1--157:38}.
\newblock


\bibitem[\protect\citeauthoryear{Huber, Hu, Tena, Mortazavian, Koppen,
  Christmas, R{\"{a}}tsch, and Kittler}{Huber et~al\mbox{.}}{2016}]%
        {Huber2015}
\bibfield{author}{\bibinfo{person}{Patrik Huber}, \bibinfo{person}{Guosheng
  Hu}, \bibinfo{person}{Rafael Tena}, \bibinfo{person}{Pouria Mortazavian},
  \bibinfo{person}{Wollem~P. Koppen}, \bibinfo{person}{William Christmas},
  \bibinfo{person}{Matthias R{\"{a}}tsch}, {and} \bibinfo{person}{Josef
  Kittler}.} \bibinfo{year}{2016}\natexlab{}.
\newblock \showarticletitle{A Multiresolution 3D Morphable Face Model and
  Fitting Framework}. In \bibinfo{booktitle}{\emph{the 11th Joint Conference on
  Computer Vision, Imaging and Computer Graphics Theory and Applications
  (VISIGRAPP)}}, Vol.~\bibinfo{volume}{4}. \bibinfo{pages}{79--86}.
\newblock


\bibitem[\protect\citeauthoryear{Jain, Thorm{\"{a}}hlen, Seidel, and
  Theobalt}{Jain et~al\mbox{.}}{2010}]%
        {Jain2010}
\bibfield{author}{\bibinfo{person}{Arjun Jain}, \bibinfo{person}{Thorsten
  Thorm{\"{a}}hlen}, \bibinfo{person}{Hans{-}Peter Seidel}, {and}
  \bibinfo{person}{Christian Theobalt}.} \bibinfo{year}{2010}\natexlab{}.
\newblock \showarticletitle{MovieReshape: tracking and reshaping of humans in
  videos}.
\newblock \bibinfo{journal}{\emph{ACM Transactions on Graphics (TOG)}}
  \bibinfo{volume}{29}, \bibinfo{number}{6} (\bibinfo{year}{2010}),
  \bibinfo{pages}{148}.
\newblock


\bibitem[\protect\citeauthoryear{Kaufmann, Wang, Sorkine{-}Hornung,
  Sorkine{-}Hornung, Smolic, and Gross}{Kaufmann et~al\mbox{.}}{2013}]%
        {Kaufmann2013}
\bibfield{author}{\bibinfo{person}{Peter Kaufmann}, \bibinfo{person}{Oliver
  Wang}, \bibinfo{person}{Alexander Sorkine{-}Hornung}, \bibinfo{person}{Olga
  Sorkine{-}Hornung}, \bibinfo{person}{Aljoscha Smolic}, {and}
  \bibinfo{person}{Markus~H. Gross}.} \bibinfo{year}{2013}\natexlab{}.
\newblock \showarticletitle{Finite Element Image Warping}.
\newblock \bibinfo{journal}{\emph{Computer Graphics Forum}}
  \bibinfo{volume}{32}, \bibinfo{number}{2} (\bibinfo{year}{2013}),
  \bibinfo{pages}{31--39}.
\newblock


\bibitem[\protect\citeauthoryear{Kim, Woo, Lee, and Kweon}{Kim
  et~al\mbox{.}}{2019}]%
        {kim2019deep}
\bibfield{author}{\bibinfo{person}{Dahun Kim}, \bibinfo{person}{Sanghyun Woo},
  \bibinfo{person}{Joon{-}Young Lee}, {and} \bibinfo{person}{In~So Kweon}.}
  \bibinfo{year}{2019}\natexlab{}.
\newblock \showarticletitle{Deep Video Inpainting}. In
  \bibinfo{booktitle}{\emph{{IEEE} Conference on Computer Vision and Pattern
  Recognition (CVPR)}}. \bibinfo{pages}{5792--5801}.
\newblock


\bibitem[\protect\citeauthoryear{Lewis, Anjyo, Rhee, Zhang, Pighin, and
  Deng}{Lewis et~al\mbox{.}}{2014}]%
        {lewis2014practice}
\bibfield{author}{\bibinfo{person}{John~P. Lewis}, \bibinfo{person}{Ken Anjyo},
  \bibinfo{person}{Taehyun Rhee}, \bibinfo{person}{Mengjie Zhang},
  \bibinfo{person}{Fr{\'{e}}d{\'{e}}ric~H. Pighin}, {and}
  \bibinfo{person}{Zhigang Deng}.} \bibinfo{year}{2014}\natexlab{}.
\newblock \showarticletitle{Practice and Theory of Blendshape Facial Models}.
  In \bibinfo{booktitle}{\emph{the 35th Annual Conference of the European
  Association for Computer Graphics}}. \bibinfo{pages}{199--218}.
\newblock


\bibitem[\protect\citeauthoryear{Leyvand, Cohen{-}Or, Dror, and
  Lischinski}{Leyvand et~al\mbox{.}}{2008}]%
        {leyvand2008data}
\bibfield{author}{\bibinfo{person}{Tommer Leyvand}, \bibinfo{person}{Daniel
  Cohen{-}Or}, \bibinfo{person}{Gideon Dror}, {and} \bibinfo{person}{Dani
  Lischinski}.} \bibinfo{year}{2008}\natexlab{}.
\newblock \showarticletitle{Data-Driven Enhancement of Facial Attractiveness}.
\newblock \bibinfo{journal}{\emph{ACM Transactions on Graphics (TOG)}}
  \bibinfo{volume}{27}, \bibinfo{number}{3} (\bibinfo{year}{2008}),
  \bibinfo{pages}{38}.
\newblock


\bibitem[\protect\citeauthoryear{Li, Bladin, Zhao, Chinara, Ingraham, Xiang,
  Ren, Prasad, Kishore, Xing, and Li}{Li et~al\mbox{.}}{2020}]%
        {li_learning_2020}
\bibfield{author}{\bibinfo{person}{Ruilong Li}, \bibinfo{person}{Karl Bladin},
  \bibinfo{person}{Yajie Zhao}, \bibinfo{person}{Chinmay Chinara},
  \bibinfo{person}{Owen Ingraham}, \bibinfo{person}{Pengda Xiang},
  \bibinfo{person}{Xinglei Ren}, \bibinfo{person}{Pratusha Prasad},
  \bibinfo{person}{Bipin Kishore}, \bibinfo{person}{Jun Xing}, {and}
  \bibinfo{person}{Hao Li}.} \bibinfo{year}{2020}\natexlab{}.
\newblock \showarticletitle{Learning {Formation} of {Physically}-{Based} {Face}
  {Attributes}}. In \bibinfo{booktitle}{\emph{2020 {IEEE}/{CVF} {Conference} on
  {Computer} {Vision} and {Pattern} {Recognition} ({CVPR})}}.
  \bibinfo{publisher}{IEEE}, \bibinfo{address}{Seattle, WA, USA},
  \bibinfo{pages}{3407--3416}.
\newblock


\bibitem[\protect\citeauthoryear{Nagano, Seo, Xing, Wei, Li, Saito, Agarwal,
  Fursund, and Li}{Nagano et~al\mbox{.}}{2019}]%
        {nagano_pagan_2019}
\bibfield{author}{\bibinfo{person}{Koki Nagano}, \bibinfo{person}{Jaewoo Seo},
  \bibinfo{person}{Jun Xing}, \bibinfo{person}{Lingyu Wei},
  \bibinfo{person}{Zimo Li}, \bibinfo{person}{Shunsuke Saito},
  \bibinfo{person}{Aviral Agarwal}, \bibinfo{person}{Jens Fursund}, {and}
  \bibinfo{person}{Hao Li}.} \bibinfo{year}{2019}\natexlab{}.
\newblock \showarticletitle{{paGAN}: real-time avatars using dynamic textures}.
\newblock \bibinfo{journal}{\emph{ACM Transactions on Graphics}}
  \bibinfo{volume}{37} (\bibinfo{year}{2019}), \bibinfo{pages}{1--12}.
\newblock


\bibitem[\protect\citeauthoryear{Ren, Yu, Zhang, Li, Liu, and Li}{Ren
  et~al\mbox{.}}{2019}]%
        {ren2019structureflow}
\bibfield{author}{\bibinfo{person}{Yurui Ren}, \bibinfo{person}{Xiaoming Yu},
  \bibinfo{person}{Ruonan Zhang}, \bibinfo{person}{Thomas~H. Li},
  \bibinfo{person}{Shan Liu}, {and} \bibinfo{person}{Ge Li}.}
  \bibinfo{year}{2019}\natexlab{}.
\newblock \showarticletitle{StructureFlow: Image Inpainting via Structure-Aware
  Appearance Flow}. In \bibinfo{booktitle}{\emph{{IEEE/CVF} International
  Conference on Computer Vision (ICCV)}}. \bibinfo{pages}{181--190}.
\newblock


\bibitem[\protect\citeauthoryear{Schaefer, McPhail, and Warren}{Schaefer
  et~al\mbox{.}}{2006}]%
        {schaefer2006image}
\bibfield{author}{\bibinfo{person}{Scott Schaefer}, \bibinfo{person}{Travis
  McPhail}, {and} \bibinfo{person}{Joe~D. Warren}.}
  \bibinfo{year}{2006}\natexlab{}.
\newblock \showarticletitle{Image deformation using moving least squares}.
\newblock Vol.~\bibinfo{volume}{25}. \bibinfo{pages}{533--540}.
\newblock


\bibitem[\protect\citeauthoryear{Shih, Lai, and Liang}{Shih
  et~al\mbox{.}}{2019}]%
        {shih2019distortion}
\bibfield{author}{\bibinfo{person}{Yi{-}Chang Shih},
  \bibinfo{person}{Wei{-}Sheng Lai}, {and} \bibinfo{person}{Chia{-}Kai Liang}.}
  \bibinfo{year}{2019}\natexlab{}.
\newblock \showarticletitle{Distortion-Free Wide-Angle Portraits on Camera
  Phones}.
\newblock \bibinfo{journal}{\emph{ACM Transactions on Graphics (TOG)}}
  \bibinfo{volume}{38}, \bibinfo{number}{4} (\bibinfo{year}{2019}),
  \bibinfo{pages}{61:1--61:12}.
\newblock


\bibitem[\protect\citeauthoryear{Shih, Paris, Barnes, Freeman, and Durand}{Shih
  et~al\mbox{.}}{2014}]%
        {shih10.1145/2601097.2601137}
\bibfield{author}{\bibinfo{person}{Yi{-}Chang Shih}, \bibinfo{person}{Sylvain
  Paris}, \bibinfo{person}{Connelly Barnes}, \bibinfo{person}{William~T.
  Freeman}, {and} \bibinfo{person}{Fr{\'{e}}do Durand}.}
  \bibinfo{year}{2014}\natexlab{}.
\newblock \showarticletitle{Style Transfer for Headshot Portraits}.
\newblock \bibinfo{journal}{\emph{ACM Transactions on Graphics (TOG)}}
  \bibinfo{volume}{33}, \bibinfo{number}{4} (\bibinfo{year}{2014}),
  \bibinfo{pages}{148:1--148:14}.
\newblock


\bibitem[\protect\citeauthoryear{Tewari, Elgharib, R, Bernard, Seidel, Pérez,
  Zollhöfer, and Theobalt}{Tewari et~al\mbox{.}}{2020}]%
        {tewari_pie_2020}
\bibfield{author}{\bibinfo{person}{Ayush Tewari}, \bibinfo{person}{Mohamed
  Elgharib}, \bibinfo{person}{Mallikarjun~B R}, \bibinfo{person}{Florian
  Bernard}, \bibinfo{person}{Hans-Peter Seidel}, \bibinfo{person}{Patrick
  Pérez}, \bibinfo{person}{Michael Zollhöfer}, {and}
  \bibinfo{person}{Christian Theobalt}.} \bibinfo{year}{2020}\natexlab{}.
\newblock \showarticletitle{{PIE}: portrait image embedding for semantic
  control}.
\newblock \bibinfo{journal}{\emph{ACM Transactions on Graphics}}
  \bibinfo{volume}{39}, \bibinfo{number}{6} (\bibinfo{year}{2020}),
  \bibinfo{pages}{1--14}.
\newblock


\bibitem[\protect\citeauthoryear{Thies, Zollh{\"{o}}fer, Nie{\ss}ner,
  Valgaerts, Stamminger, and Theobalt}{Thies et~al\mbox{.}}{2015}]%
        {thies2015expression_transfer}
\bibfield{author}{\bibinfo{person}{Justus Thies}, \bibinfo{person}{Michael
  Zollh{\"{o}}fer}, \bibinfo{person}{Matthias Nie{\ss}ner},
  \bibinfo{person}{Levi Valgaerts}, \bibinfo{person}{Marc Stamminger}, {and}
  \bibinfo{person}{Christian Theobalt}.} \bibinfo{year}{2015}\natexlab{}.
\newblock \showarticletitle{Real-time Expression Transfer for Facial
  Reenactment}.
\newblock \bibinfo{journal}{\emph{ACM Transactions on Graphics (TOG)}}
  \bibinfo{volume}{34}, \bibinfo{number}{6} (\bibinfo{year}{2015}),
  \bibinfo{pages}{183:1--183:14}.
\newblock


\bibitem[\protect\citeauthoryear{Thies, Zollh{\"{o}}fer, Stamminger, Theobalt,
  and Nie{\ss}ner}{Thies et~al\mbox{.}}{2016}]%
        {thies2016face2face}
\bibfield{author}{\bibinfo{person}{Justus Thies}, \bibinfo{person}{Michael
  Zollh{\"{o}}fer}, \bibinfo{person}{Marc Stamminger},
  \bibinfo{person}{Christian Theobalt}, {and} \bibinfo{person}{Matthias
  Nie{\ss}ner}.} \bibinfo{year}{2016}\natexlab{}.
\newblock \showarticletitle{Face2Face: Real-Time Face Capture and Reenactment
  of {RGB} Videos}. In \bibinfo{booktitle}{\emph{{IEEE} Conference on Computer
  Vision and Pattern Recognition (CVPR)}}. \bibinfo{pages}{2387--2395}.
\newblock


\bibitem[\protect\citeauthoryear{Tran, Liu, and Liu}{Tran
  et~al\mbox{.}}{2019}]%
        {Trana}
\bibfield{author}{\bibinfo{person}{Luan Tran}, \bibinfo{person}{Feng Liu},
  {and} \bibinfo{person}{Xiaoming Liu}.} \bibinfo{year}{2019}\natexlab{}.
\newblock \showarticletitle{Towards High-Fidelity Nonlinear 3D Face Morphable
  Model}. In \bibinfo{booktitle}{\emph{{IEEE} Conference on Computer Vision and
  Pattern Recognition (CVPR)}}. \bibinfo{pages}{1126--1135}.
\newblock


\bibitem[\protect\citeauthoryear{Tran and Liu}{Tran and Liu}{2018}]%
        {Tran}
\bibfield{author}{\bibinfo{person}{Luan Tran} {and} \bibinfo{person}{Xiaoming
  Liu}.} \bibinfo{year}{2018}\natexlab{}.
\newblock \showarticletitle{Nonlinear 3D Face Morphable Model}. In
  \bibinfo{booktitle}{\emph{{IEEE} Conference on Computer Vision and Pattern
  Recognition (CVPR)}}. \bibinfo{pages}{7346--7355}.
\newblock


\bibitem[\protect\citeauthoryear{Triggs, McLauchlan, Hartley, and
  Fitzgibbon}{Triggs et~al\mbox{.}}{1999}]%
        {triggs1999bundle}
\bibfield{author}{\bibinfo{person}{Bill Triggs}, \bibinfo{person}{Philip~F.
  McLauchlan}, \bibinfo{person}{Richard~I. Hartley}, {and}
  \bibinfo{person}{Andrew~W. Fitzgibbon}.} \bibinfo{year}{1999}\natexlab{}.
\newblock \showarticletitle{Bundle Adjustment - {A} Modern Synthesis}. In
  \bibinfo{booktitle}{\emph{International Workshop on Vision Algorithms}},
  Vol.~\bibinfo{volume}{1883}. \bibinfo{pages}{298--372}.
\newblock


\bibitem[\protect\citeauthoryear{Xiao, Tang, Wu, Jin, Yang, and Jin}{Xiao
  et~al\mbox{.}}{2020}]%
        {XiaoPortraits}
\bibfield{author}{\bibinfo{person}{Qinjie Xiao}, \bibinfo{person}{Xiangjun
  Tang}, \bibinfo{person}{You Wu}, \bibinfo{person}{Leyang Jin},
  \bibinfo{person}{Yong{-}Liang Yang}, {and} \bibinfo{person}{Xiaogang Jin}.}
  \bibinfo{year}{2020}\natexlab{}.
\newblock \showarticletitle{Deep Shapely Portraits}. In
  \bibinfo{booktitle}{\emph{{MM} '20: The 28th {ACM} International Conference
  on Multimedia}}. \bibinfo{pages}{1800--1808}.
\newblock


\bibitem[\protect\citeauthoryear{Zeng, Liu, Lin, and Ge}{Zeng
  et~al\mbox{.}}{2020}]%
        {ZengTalking}
\bibfield{author}{\bibinfo{person}{Dan Zeng}, \bibinfo{person}{Han Liu},
  \bibinfo{person}{Hui Lin}, {and} \bibinfo{person}{Shiming Ge}.}
  \bibinfo{year}{2020}\natexlab{}.
\newblock \showarticletitle{Talking Face Generation with Expression-Tailored
  Generative Adversarial Network}. In \bibinfo{booktitle}{\emph{Proceedings of
  the 28th ACM International Conference on Multimedia}}
  \emph{(\bibinfo{series}{MM '20})}. \bibinfo{pages}{1716–1724}.
\newblock


\bibitem[\protect\citeauthoryear{Zhao, Jin, Huang, Chai, and Zhou}{Zhao
  et~al\mbox{.}}{2018}]%
        {Zhao2018Portrait}
\bibfield{author}{\bibinfo{person}{Haiming Zhao}, \bibinfo{person}{Xiaogang
  Jin}, \bibinfo{person}{Xiaojian Huang}, \bibinfo{person}{Menglei Chai}, {and}
  \bibinfo{person}{Kun Zhou}.} \bibinfo{year}{2018}\natexlab{}.
\newblock \showarticletitle{Parametric Reshaping of Portrait Images for
  Weight-change}.
\newblock \bibinfo{journal}{\emph{{IEEE} Computer Graphics and Applications}}
  \bibinfo{volume}{38}, \bibinfo{number}{1} (\bibinfo{year}{2018}),
  \bibinfo{pages}{77--90}.
\newblock


\bibitem[\protect\citeauthoryear{Zollh{\"{o}}fer, Thies, Garrido, Bradley,
  Beeler, P{\'{e}}rez, Stamminger, Nie{\ss}ner, and Theobalt}{Zollh{\"{o}}fer
  et~al\mbox{.}}{2018}]%
        {Thies2018}
\bibfield{author}{\bibinfo{person}{Michael Zollh{\"{o}}fer},
  \bibinfo{person}{Justus Thies}, \bibinfo{person}{Pablo Garrido},
  \bibinfo{person}{Derek Bradley}, \bibinfo{person}{Thabo Beeler},
  \bibinfo{person}{Patrick P{\'{e}}rez}, \bibinfo{person}{Marc Stamminger},
  \bibinfo{person}{Matthias Nie{\ss}ner}, {and} \bibinfo{person}{Christian
  Theobalt}.} \bibinfo{year}{2018}\natexlab{}.
\newblock \showarticletitle{State of the Art on Monocular 3D Face
  Reconstruction, Tracking, and Applications}.
\newblock \bibinfo{journal}{\emph{Computer Graphics Forum}}
  \bibinfo{volume}{37}, \bibinfo{number}{2} (\bibinfo{year}{2018}),
  \bibinfo{pages}{523--550}.
\newblock


\end{thebibliography}


\end{document}




\appendix

\section{Appendix}


 
 \subsection{Sparse control points warping}\label{app:sparsewarping}
 
 Figure 10(b) shows the case of only using a set of sparse grid points to represent the face boundary. In order to obtain reasonable results, besides the line-bending energy term $E_l$ and regularization energy term $E_r$ in Section~5.2, a distortion term $E_d$ is further added. This energy term makes the triangle formed by the contour points and their surrounding points as identical as possible in shape and area with respect to the corresponding triangle formed by the grid points before reshaping, thus makes the boundary to be more coherent.
Let $\Gamma$ and $\gamma$ be the greatest and the least singular value of the triangle deformation matrix, respectively. 
The energy term proposed first by Sander et al. [19] can be described as:
\begin{equation}
E_d=||\Gamma|| + \sqrt{\frac{1}{2}(\Gamma^2+\gamma^2)}.
\end{equation}
This equation requires that the triangular area and shape do not change significantly. To further constrain the rotation and shape change of the triangle, we add a similarity energy term $E_s$ [6] as:
\begin{equation}
E_s=\sum_{e\in M_u} ||(\bold{v}_0'-\bold{v}_1') - \frac{||\bold{v}_0-\bold{v}_1||}{||\bold{v}_2-\bold{v}_1||}\bold{R}_{\theta}(\bold{v}_2'-\bold{v}_1')||^2,
\end{equation}
where $e = (\bold{v}_0,\bold{v}_1,\bold{v}_2)$ is the triangle in the grid $M_u$ before reshaping, and $e' = (\bold{v}_0',\bold{v}_1',\bold{v}_2')$ is the corresponding triangle of $e$ in the grid $M_u$ with $\bold{R}_\theta$ being the $2 \times 2$ rotation matrix which rotates the edge ($\bold{v}_2-\bold{v}_1$) to ($\bold{v}_0-\bold{v}_1$).
In this case, we performed a two-step optimization of the face grid. The first step uses only the same energy function as Eqn. [17], which makes the deformed grid after reshaping uniform and regular. The second optimization adds the distortion term and similarity energy term with the following energy function:
\begin{equation}
E = w_lE_l+w_rE_r+w_dE_d + w_sE_s.
\end{equation}
 
 \subsection{List of video sources}
 The used video sources are listed in Table~\ref{tab:video}.

\begin{table}[]
\caption{The list of video sources\label{tab:video}}
\begin{tabular}{|l|}

\hline

{ Jennifer Lawrence Shares Her Most Embarrassing Moments} \\
\href{url}{https://www.youtube.com/watch?v=4OTQV48qBoY      }                             \\ \hline

President Barack Obama: Between Two Ferns with Zach Galifianakis              \\
\href{url}{https://www.youtube.com/watch?v=UnW3xkHxIEQ  }                                 \\ \hline

{ Moon Jae-in and Kim Jong-un address 150,000 crowd | Al Jazeera English} \\
\href{url}{ https://www.youtube.com/watch?v=NXyQSvbKGIM  }                              \\ \hline

{ Mark Zuckerberg : How to Build the Future} \\
\href{url}{https://www.youtube.com/watch?v=Lb4IcGF5iTQ }                                 \\ \hline

{ How To Flirt Via TEXT Message | 10 Texting Tips} \\
\href{url}{https://www.youtube.com/watch?v=fWVqk7FTcnw   }                               \\ \hline

{ Sam Altman : How to Build the Future} \\
\href{url}{https://www.youtube.com/watch?v=sYMqVwsewSg   }                               \\ \hline

{ Broadsheet Owner’s Exclusive Interview | Why General Malek Wanted Corrupt Commission From Him} \\
\href{url}{https://www.youtube.com/watch?v=c55-hlzbc8k   }                               \\ \hline

{ WATCH: Joe Biden gives first speech as president} \\
\href{url}{https://www.youtube.com/watch?v=cTtKDN4LgL8  }                                \\ \hline

{Putin wishes Russians brighter New Year, 'return to normal'} \\
\href{url}{https://www.youtube.com/watch?v=1kn5JBEHzD4    } 
\\ \hline

{5 Simple Journalist Techniques for Effective Interviews} \\
\href{url}{https://www.youtube.com/watch?v=NWDL\_UYMc7Q}                           
\\ \hline
{To Be Honest 2.0 | Mathira| Tabish Hashmi | Full Episode | NashpatiPrime} \\
\href{url}{https://www.youtube.com/watch?v=nhMaY17m11w}                           
\\ \hline
{Elon Musk : How to Build the Future} \\
\href{url}{https://www.youtube.com/watch?v=tnBQmEqBCY0}                           
\\ \hline

{Dwight's Acceptance Speech -The Office} \\
\href{url}{https://www.youtube.com/watch?v=ilE4lr9Qb3A}                           
\\ \hline

\end{tabular}
\end{table}